\documentclass[journal]{IEEEtran}

\usepackage{cite}
\ifCLASSINFOpdf
  \usepackage[pdftex]{graphicx}
  \graphicspath{{../pdf/}{../jpeg/}}
  \DeclareGraphicsExtensions{.pdf,.jpeg,.png}
\else
  \usepackage[dvips]{graphicx}
  \graphicspath{{../eps/}}
  \DeclareGraphicsExtensions{.eps}
\fi
\usepackage{amsmath}
\usepackage{amssymb}
\usepackage{amsfonts}
\usepackage{multirow}
\usepackage{booktabs}
\usepackage{tabularx}
\usepackage{subfloat}
\usepackage{epstopdf}
\usepackage[labelfont=bf,textfont=bf]{caption}
\usepackage{algorithmic}
\usepackage{array}
\usepackage[caption=false,font=normalsize,labelfont=sf,textfont=sf]{subfig}
\usepackage{fixltx2e}
\usepackage{stfloats}
\usepackage{url}
\begin{document}
%\title{Exploiting Spatio-Temporal Relation via Local-Global Transformer for Text-Video Retrieval}

\title{BiC-Net: Learning Efficient Spatio-Temporal Relation for Text-Video Retrieval}


\author{Ning~Han, Jingjing~Chen, Chuhao Shi, Yawen Zeng, Guangyi~Xiao, and~Hao~Chen % <-this % stops a space
 
 %\thanks{This work was partially supported by the National Key Research and Development Project of China (2018YFB1402600) and the National Natural Science Foundation of China (61772190). (Corresponding author: JingJing Chen and Hao Chen.)}

\thanks{Ning Han, Chuhao~Shi, Guangyi~Xiao, and Hao Chen are with the Department of Information Science and Engineering, Hunan University, Changsha (e-mail: ninghan@hnu.edu.cn; sch8288@hnu.edu.cn; guangyi.xiao@gmail.com; chenhao@hnu.edu.cn).}% <-this % stops a space

\thanks{Jingjing Chen is with the Department of Computer Science, Fudan University. Shanghai (e-mail: chenjingjing@fudan.edu.cn).}

\thanks{Yawen Zeng is with the Bytedance AI Lab, Beijing (e-mail: yawenzeng11@gmail.com).}

} % <-this % stops a space

 %%\thanks{(Corresponding author: Guangyi Xiao.)}

%%IEEE TRANSACTIONS ON MULTIMEDIA
\markboth{}%
{Shell \MakeLowercase{\textit{et al.}}: Bare Demo of IEEEtran.cls for IEEE Journals}



\maketitle

\begin{abstract}
The task of text-video retrieval aims to understand the correspondence between language and vision, has gained increasing attention in recent years. Previous studies either adopt off-the-shelf 2D/3D-CNN and then use average/max pooling to directly capture spatial features with aggregated temporal information as global video embeddings, or introduce graph-based models and expert knowledge to learn local spatial-temporal relations. However, the existing methods have two limitations: 1) The global video representations learn video temporal information in a simple average/max pooling manner and do not fully explore the temporal information between every two frames. 2) The graph-based local video representations are handcrafted, it depends heavily on expert knowledge and empirical feedback, which may not be able to effectively mine the higher-level fine-grained visual relations. These limitations result in their inability to distinguish videos with the same visual components but with different relations. 

To solve this problem, we propose a novel cross-modal retrieval framework, Bi-Branch Complementary Network (BiC-Net), which modifies transformer architecture to effectively bridge text-video modalities in a complementary manner via combining local spatial-temporal relation and global temporal information. Specifically, local video representations are encoded using multiple transformer blocks and additional residual blocks to learn spatio-temporal relation features, calling the module a Spatio-Temporal Residual transformer (SRT). Meanwhile, Global video representations are encoded using a multi-layer transformer block to learn global temporal features. Finally, we align the spatio-temporal relation and global temporal features with the text feature on two embedding spaces for cross-modal text-video retrieval. Extensive experiments are conducted on MSR-VTT, MSVD, and YouCook2 datasets. The results demonstrate the effectiveness of our proposed model. The code is available at: \emph{\url{https://github.com/lionel-hing/BiC-Net}}. 
\end{abstract}

\begin{IEEEkeywords}
Text-Video Retrieval, Spatio-Temporal Relation,  Bi-Branch Complementary Network.  
\end{IEEEkeywords}

% \fi
\IEEEpeerreviewmaketitle

\section{Introduction}
\begin{figure}[t]
   \center
   \includegraphics[width=0.5\textwidth]{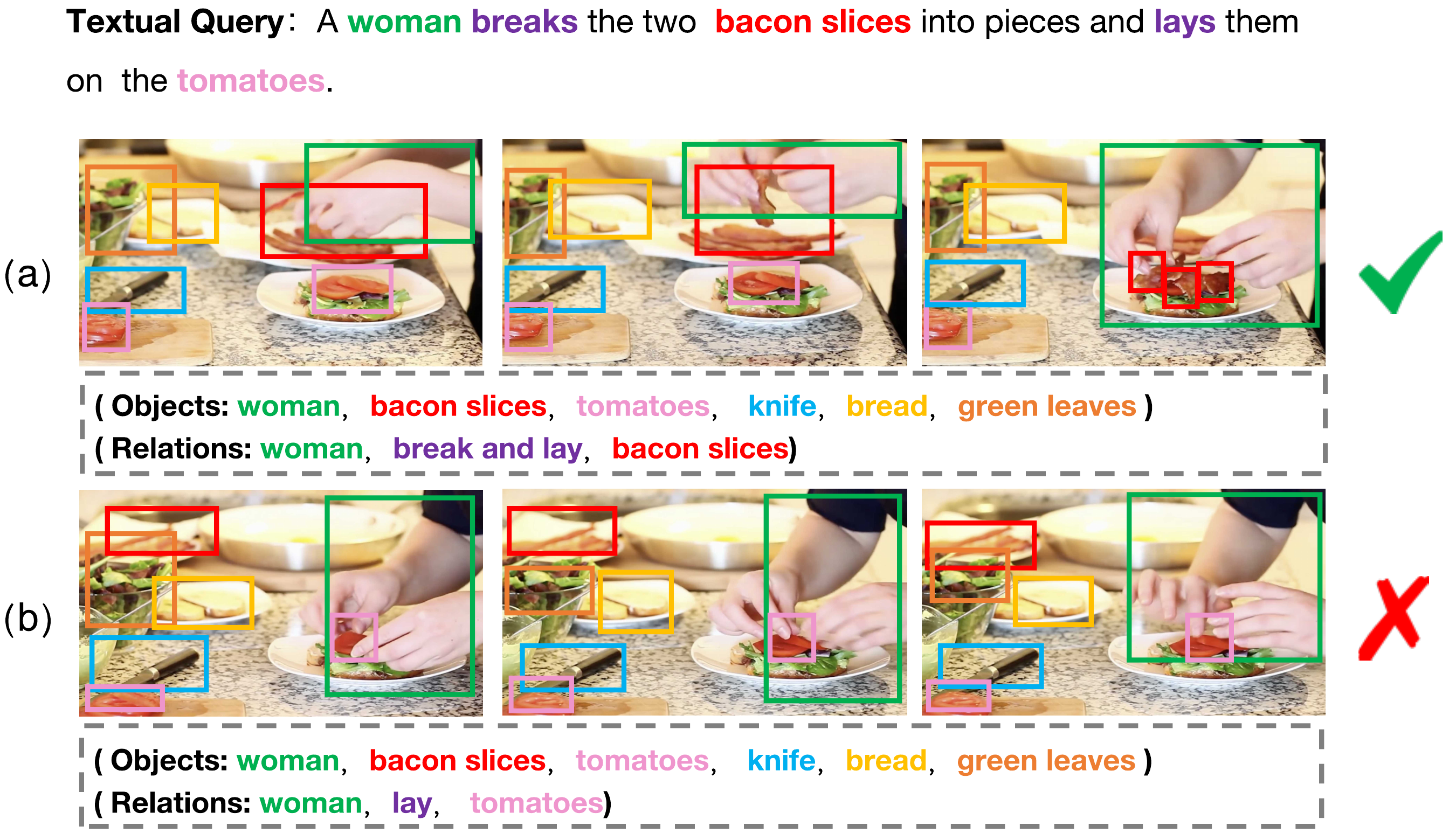}
   %\vspace{-0.1cm}
   \caption{An example of text to video retrieval. Given a textual query, a common pipeline with fine-grained \cite{song2021spatial} or global semantical visual features \cite{miech2019howto100m} will return two videos with the same compositions. The retrieval model with complementary spatio-temporal relation visual features can filter out false-positive without correct interactions.}
   \label{fig:example}
   \vspace{-0.3cm}
\end{figure}

\IEEEPARstart{R}{ecent} years have witnessed an exponential growth of multimedia data (e.g., video, image, and text), which increases the demands for effectively retrieving relevant data from another modality, when given a query of one modality. Being one of these challenging tasks, text-video retrieval aims to retrieve the video given a text query, which requires measuring the semantic similarity between a sentence and a video. Video data are distinct from images due to the temporal dependencies among frames and the additional dynamic relationships among objects, resulting in the inability of existing video retrieval techniques to distinguish videos with the same visual components but with different relations. Figure \ref{fig:example} shows such an example. Given the text query ``\emph{A woman breaks the two bacon slices into pieces and lay them on the tomatoes}'', the existing retrieval systems are likely to consider both (a) and (b) as positive examples, since both of them contain the same motion (``\emph{laying}'') and objects (``\emph{bacon slices}'', ``\emph{tomatoes}'') with the text query. However, example (b) is indeed a false positive, as it presents ``\emph{a woman lays tomatoes on the green leaves}'' (with bacon slices on the kitchen table). This example suggests that ignoring the visual relations (i.e., object relations) presented in videos could lead to inaccurate retrieval results. Therefore, capturing higher-level spatio-temporal visual relations in videos is crucial to distinguish similar videos.

This paper investigates the problem of cross-modal text-video retrieval. In the literature, many efforts have been devoted to learning better video representations, in order to improve the performance of text-video retrieval. Based on the granularity of feature representations, existing works can be roughly categorized into global and local feature-based methods. Global feature-based methods typically use global representations to represent entire video and sentence, which usually lose part of this temporal information and local details. Such approaches work well in a simple cross-modal retrieval scenario, where only a single object is presented in the video or text query. For more realistic cases involving complex natural scenes, the performance of these methods is usually unsatisfactory. In contrast, local feature-based methods pay attention to local details and perform matching by detecting objects in videos and texts. With local region modeling, the performance of text-video retrieval has been significantly improved. Nevertheless, the existing efforts can only capture simple visual relations by graph convolutional network (GCN) \cite{feng2020exploiting, song2021spatial} or utilize an attention mechanism \cite{chen2020fine, wu2021hanet} as a cross-modal interaction module to delve into high-level correspondences. As GCN-based video modeling is handcrafted, it depends heavily on expert knowledge and empirical feedback, which may not be able to effectively mine and model the higher-level fine-grained visual relations. Attention-based models, on the other hand, selectively align the key information presented in different modalities. As the fine-grained visual relations are also ignored by attention-based methods, the performance of these methods is still unsatisfactory, and novel modeling solutions are eagerly awaited.

To further improve the performance of text-video retrieval, this paper studies this problem from the perspective of spatio-temporal relation modeling for videos. Generally, there are two major obstacles in modeling the spatio-temporal relation. First, videos contain diverse spatial and temporal information within variations in motion and richer information in local visual details. These objects and interactions increase the difficulties in capturing higher-level fine-grained visual contents. Second, local relation modeling captures considerable fragmented information, which will overlook contextual information. Therefore, the way to comprehensively capture multi-granularity visual information to represent videos from complementary spatial and temporal perspectives is of great importance.

\begin{figure*}[t]
   \center
   \includegraphics[width=0.99\textwidth]{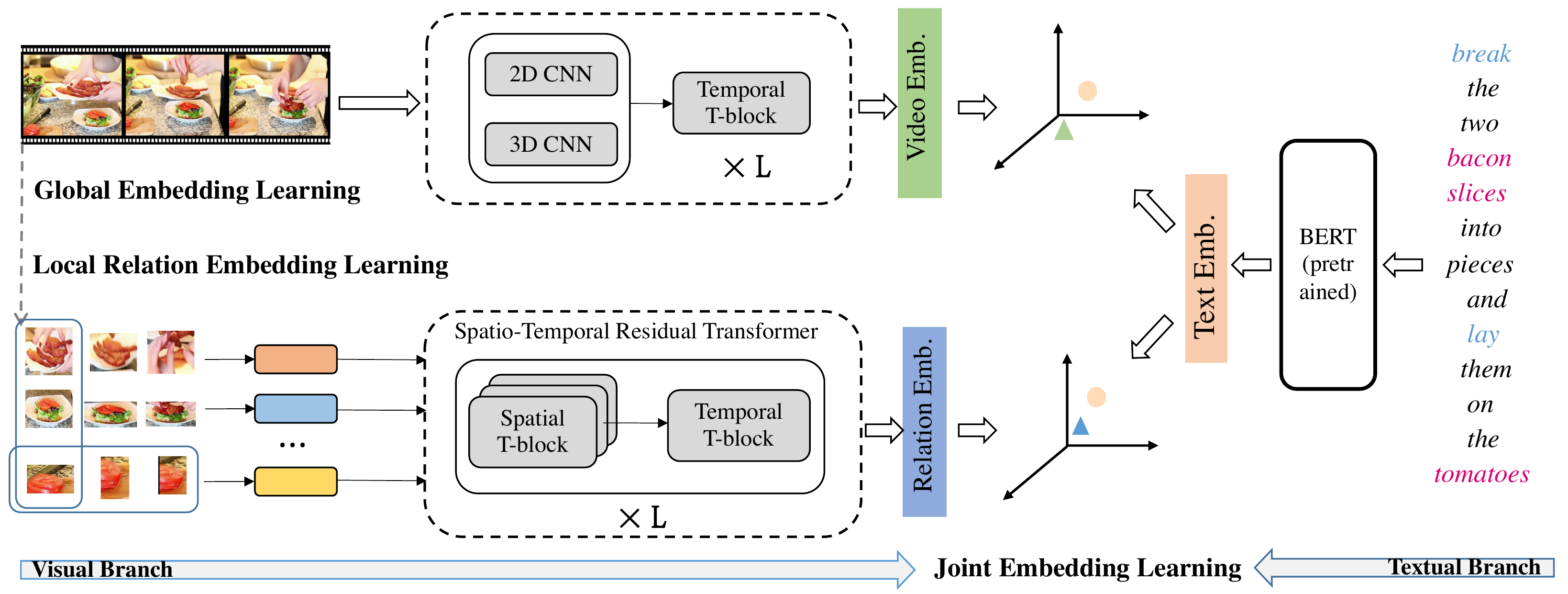}
   %\vspace{-0.1cm}
   \caption{The overall framework of our proposed BiC-Net. First, we extract local relational and global visual features for videos. The local relations are represented by local regional features using a spatio-temporal residual transformer. The global video features are represented by 2D-CNN and 3D-CNN features via a multi-layer temporal transformer. Then, we extract textual features by BERT. Finally, both video and relational features are leveraged to align with textual features on two embedding spaces for cross-modal text-video retrieval. Among them, T-Block denotes the transformer block.}
   \label{fig:framework}
   %\vspace{-0.2cm}
\end{figure*}

To address the aforementioned problems, we propose a novel Bi-Branch Complementary Network (BiC-Net), which modifies transformer architecture to effectively bridge text-video modalities in a complementary manner via combining local spatial-temporal relation and global temporal information. We present an overview of BiC-Net in Figure \ref{fig:framework}. Specifically, for videos, our BiC-Net attempts to extract two perspectives of features — global temporal features and local relation features. At the global temporal level, we directly adopt the widely used 2D and 3D-CNN. For local relational features, we use pre-trained Faster-RCNN \cite{ren2015faster} to extract regional features (i.e., features of bounding boxes). Then, a spatio-temporal residual transformer is employed for learning high-level fine-grained relational features. This module separately captures local spatial relations, and long-term temporal relations among local spatial relations. In addition, a multi-layer transformer block is applied for learning global temporal features. To cover different levels of semantics, we align the global temporal and local relation features with the text feature on two embedding spaces. Lastly, the similarity between videos and texts is measured in both embedding spaces and then summed to obtain the final similarity score. In this way, the global temporal information and local relation information in a video can be utilized for cross-modal text-video retrieval comprehensively.

Our contributions are summarized as below:

\begin{itemize}

\item We incorporate feature-split with bi-branch framework called BiC-Net to capture local relations and global temporal features comprehensively, which aligns the global temporal and local relation features with the text feature on two embedding spaces for cross-modal text-video retrieval.
 
\item We first introduce a simple and effective spatio-temporal residual transformer to learn higher-level local relation features, and a multi-layer temporal transformer to further explore global temporal information for global temporal features. In this way, the bi-branch information in video and text can accurately capture cross-modal semantic alignment in a cooperative and complementary manner.

\item We conduct extensive experiments on three standard benchmarks and verify the effectiveness of our proposed method by showing that BiC-Net can achieve SOTA performance (86.7\% on MSR-VTT 1k-A test set) under similar conditions.

\end{itemize}

The rest of this paper is organized as follows. In Section \ref{sec:rw}, we briefly describe a review of related work. In Section \ref{sec:method}, we describe our proposed BiC-Net model. In Section \ref{sec:exp}, we provide implementation details and experimental results. In Section \ref{sec:con}, we finally conclude our paper.

%-------------------------------------------------------------------------

\section{Related Work}\label{sec:rw}

\subsection{Text-video Retrieval}\label{sec:tvr}

According to the granularity of feature representations, we roughly divide existing works into two groups: global feature-based methods and local relation feature-based methods.

\textbf{Global feature-based methods} \cite{miech2019howto100m, yang2020tree, dong2021dual} extract global feature representations of videos and texts and then learn a joint embedding space where visual and textual similarity is measured. For the video representation, they adopt 2D/3D CNN models to extract frame features and aggregate frame features by average-pooling \cite{dong2018predicting, liu2019use, li2020sea} or max-pooling \cite{miech2018learning, miech2019howto100m}. For the video representation, they focused only on leveraging the global feature of the video. For instance, Dong et al. \cite {dong2019dual, dong2021dual} employ three levels, i.e., global, temporal, and local to encode videos and texts and learn a hybrid common space for video-text similarity measurement. Miech et al. \cite{miech2019howto100m} adopt 2D and 3D CNN to extract frame features and only use max-pooling to obtain global video representation. Yang et al. \cite{yang2020tree} present a latent semantic tree to encode the text and used a multi-head self-attention mechanism to obtain the temporal-attentive video representation.

\textbf{Local feature-based methods} \cite{wray2019fine, chen2020fine, feng2020exploiting, han2021fine, wu2021hanet, han2022adversarial} use local semantic information from language or video for better text-video alignment from different aspects and then perform text-video retrieval tasks. Wray et al.\cite{wray2019fine} disentangle action phrases into verbs and nouns for fine-grained video retrieval. The graph-based approaches \cite{feng2020exploiting, song2021spatial, han2022adversarial} construct different semantic correlation graphs for videos and learn fine-grained semantic relations for text-video retrieval. Some works \cite{chen2020fine, han2021fine, wu2021hanet} also propose fine-grained alignment models that decompose text and video into multiple levels and align text with video at multiple levels for text-video matching.

Recently, some studies have also explored a combination of video experts (e.g., motion, audio, and speech) \cite{mithun2018learning, liu2019use, gabeur2020multi, wang2021t2vlad} or pre-trained video experts \cite{miech2020end, rouditchenko2020avlnet, patrick2020support} to improve the performance of cross-modal retrieval. Lately, Transformer-based works \cite{luo2021clip4clip, lei2021less, fang2021clip2video, liu2021hit, bain2021frozen} have benefited from pre-training models on large-scale language-vision datasets \cite{bain2021frozen, luo2021clip4clip}. For example, Bain et al. \cite{bain2021frozen} propose an end-to-end trainable model which adopts a space-time transformer encoder to flexibly train on both video and image datasets. Luo et al. \cite{luo2021clip4clip} apply the joint language-vision model of CLIP \cite{radford2021learning}, pre-trained on a large-scale text-image dataset as a backbone for text-video retrieval. However, Transformer-based methods have a heavy computational burden due to computational intensive operations and are extremely time-consuming to pre-train on large-scale datasets. Different from these existing works, our study introduces a new spatial-temporal residual transformer to learn higher-level local relation features and a multi-layer transformer to further explore global temporal information for global temporal features. In this way, bi-branch information in video and text can accurately capture cross-modal semantic alignment in a cooperative and complementary manner. 

\subsection{Spatio-Temporal Relation Modeling in Video Understanding}\label{sec:vrm}

For spatio-temporal relation modeling in video understanding, earlier works adopt 2D/3D CNNs to represent the core operators for spatio-temporal feature learning across downstream video tasks \cite{lu2021learning, zhang2020hybrid, qi2021semantics, wang2022many, wang2022many}. However, these video representations focus on learning spatio-temporal features from the entire video and can hardly capture local spatial-temporal relation information. To understand the local relation information in the video, several efforts have demonstrated the effectiveness of incorporating local spatial-temporal relationships into video understanding in many downstream applications, such as visual relationship detection \cite{qian2019video, xiao2020visual, li2021interventional}, action recognition \cite{wang2018videos, yan2018spatial, wu2019learning}, and video retrieval \cite{feng2020exploiting, song2021spatial, han2022adversarial}. For instance,  Qian et al. construct a spatio-temporal graph in adjacent video clips to define the relationships between objects. Wang et al.  \cite{wang2018videos} abstract the video as a space-time graphs for action recognition. Song et al. \cite{song2021spatial} model video as a spatial-temporal graph between object interactions for text-video retrieval. However, modeling object spatio-temporal relations in the video is still not thoroughly investigated. These studies have built visual relation graphs and adopted the GCN \cite{kipf2016semi} to extract visual relation graph features. Massive graph construction and graph feature extraction are hand-crafted, complex, and time-consuming. Recently, the transformer \cite{vaswani2017attention} has shown great superiority in understanding 1, 2, and 3-dimensional signals (e.g., natural language processing and computer vision), and has strong interpretability, and strong representation capabilities. Unlike these works, our work designs a spatio-temporal residual transformer to learn the local spatio-temporal relations and further mine the object interactions. Notably, we validate in the experiments that under a strict memory budget, our approach can surpass many related methods.

\section{Proposed Method}\label{sec:method}

As depicted in Figure \ref{fig:framework}, the overall pipeline of the proposed method consists of four modules: 1) video embedding learning, which involves extracting video global features; 2) relation embedding learning, which involves extracting local relational features in videos; 3) text embedding learning, which learns the representation for textual sentences by BERT \cite{devlin2019bert}; and 4) joint embedding learning, which optimizes the correspondence between text and video features in a common space with a triple ranking loss.

\subsection{Video Embedding Learning}\label{sec:riv}

Given a long video clip, we sample $T$ video frames from it with the same temporal duration between every two frames. For frame-level features, we first use 2D-CNN to extract appearance features and 3D-CNN to extract motion features. Then, we concatenate 2D and 3D features and apply a pointwise linear layer to obtain global visual features $F_{g} \in \mathbb{R}^{d_{g}}$. Finally, we feed the result to standard multi-layer transformer block \cite{vaswani2017attention} and an attention-aware feature aggregation layer \cite{ging2020coot} to obtain its video embedding, which is denoted as $F_{v} \in \mathbb{R}^{d_{*}}$.

\subsection{Relation Embedding Learning}\label{sec:rel}

In addition to having global visual features, the proposed framework learns local relation features from the video to improve the performance of cross-modal retrieval. The introduction of spatio-temporal relation among objects in the video equips the model with the ability to identify the fine-grained differences of video with similarity. To capture the visual relations from the video, we first adopt the pre-trained Faster RCNN \cite{anderson2018bottom} to detect frame-level region proposals and select the top $N$ region proposals with the highest detection confidence to represent each frame. Prior efforts \cite{wang2018videos, wu2019learning} focus on abstracting frame-level region proposals as fully connected spatial-temporal graphs and using GCN to learn relational features. However, computing all pair-wise relations across all video frames would be inefficient in creating a video as a fully connected graph. In recent years, pure transformer-based models have shown promising performance due to their strong representation capabilities. As a central piece of transformer, self-attention comes with a flexible mechanism to deal with variable-length inputs. It can be understood as a fully connected layer where the weights are dynamically generated from pairwise relations from inputs, which conveys refreshing solutions to process visual relations. 

Inspired by these pioneering efforts, to capture higher-level visual relations from the video, we design a new architecture to learn the relation embeddings, named Spatio-Temporal Residual Transformer (SRT), that exploits all the variants of transformer blocks and residual connections but composes each in different placement of SRT. In the following, the basic components used in the transformer block and the transformer block used in the SRT module are presented in detail.

%The transformer can be viewed as a GCN defined over a complete directed graph (with self-loop) where each input is a node in the graph.

\textbf{Transformer Block.} The Transformer consists of multi-head self-attention (MSA), multi-layer perceptron (MLP), and layer-norm (LN). In the self-attention module, the inputs $ X \in \mathbb{R}^{n \times d}$ are linearly transformed to three parts, i.e., queries $ Q \in \mathbb{R}^{n \times d_{k}}$, keys $ K \in \mathbb{R}^{n \times d_{k}}$ and values $ V \in \mathbb{R}^{n \times d_{v}}$, where $n$ is the sequence length, $d$, $d_{k}$, $d_{v}$ are the dimensions of inputs, queries (keys) and values, respectively. The scaled dot-product attention is applied on $Q$, $K$, $V$ :
\begin{equation}
SA \left(Q, K, V \right) = softmax \left (\frac{QK^{T}}{\sqrt{d_{k}}} \right)V.
\end{equation}
With $SA \left (Q, K, V\right)$, $MSA$ is defined as:
\begin{equation}
\begin{aligned}
MSA \left(Q, K, V \right) = Concat \left (head_{1}, \cdot \cdot \cdot, head_{M} \right)W^{O}, \\ where \quad head_{i} = SA(QW^{Q}_{i}, KW^{K}_{i}, VW^{V}_{i}).
\end{aligned}
\end{equation} 
 Where $QW^{Q}_{i}, KW^{K}_{i}, VW^{V}_{i}$ are projections of different heads, $W^{O}$ is another mapping function. The MLP is applied between self-attention layers for feature transformation and non-linearity:
\begin{equation}
MLP \left(X \right) = GELU \left(X W_{1} +b_{1} \right) W_{2} +b_{2},
\end{equation}
where $W_{1} \in \mathbb{R}^{d \times d_{m}}$ and $W_{2} \in \mathbb{R}^{d_{m} \times d}$ are weights of the two fully-connected layers respectively, $b_{1} \in \mathbb{R}^{d_{m}}$ and $b_{2} \in \mathbb{R}^{d}$ are the bias terms, and GELU \cite{hendrycks2016gaussian} is the activation function. Layer normalization \cite{ba2016layer} is a key part in transformer for stable training and faster convergence, and LN is applied over each sample $x \in \mathbb{R}^{d}$ as follows:
\begin{equation}
LN \left(x \right) =  \frac{x-\mu}{\eta} \odot \gamma + \beta,
\end{equation}
where $\mu \in \mathbb{R}$,  $\eta \in \mathbb{R}$ are the mean and standard deviation of the feature respectively, $\odot$ is the element-wise dot, and $\gamma \in \mathbb{R}^{d}$, $\beta \in \mathbb{R}^{d}$ are learnable affine transform parameters.

\begin{figure*}[t]
   \center
   \includegraphics[width=1\textwidth]{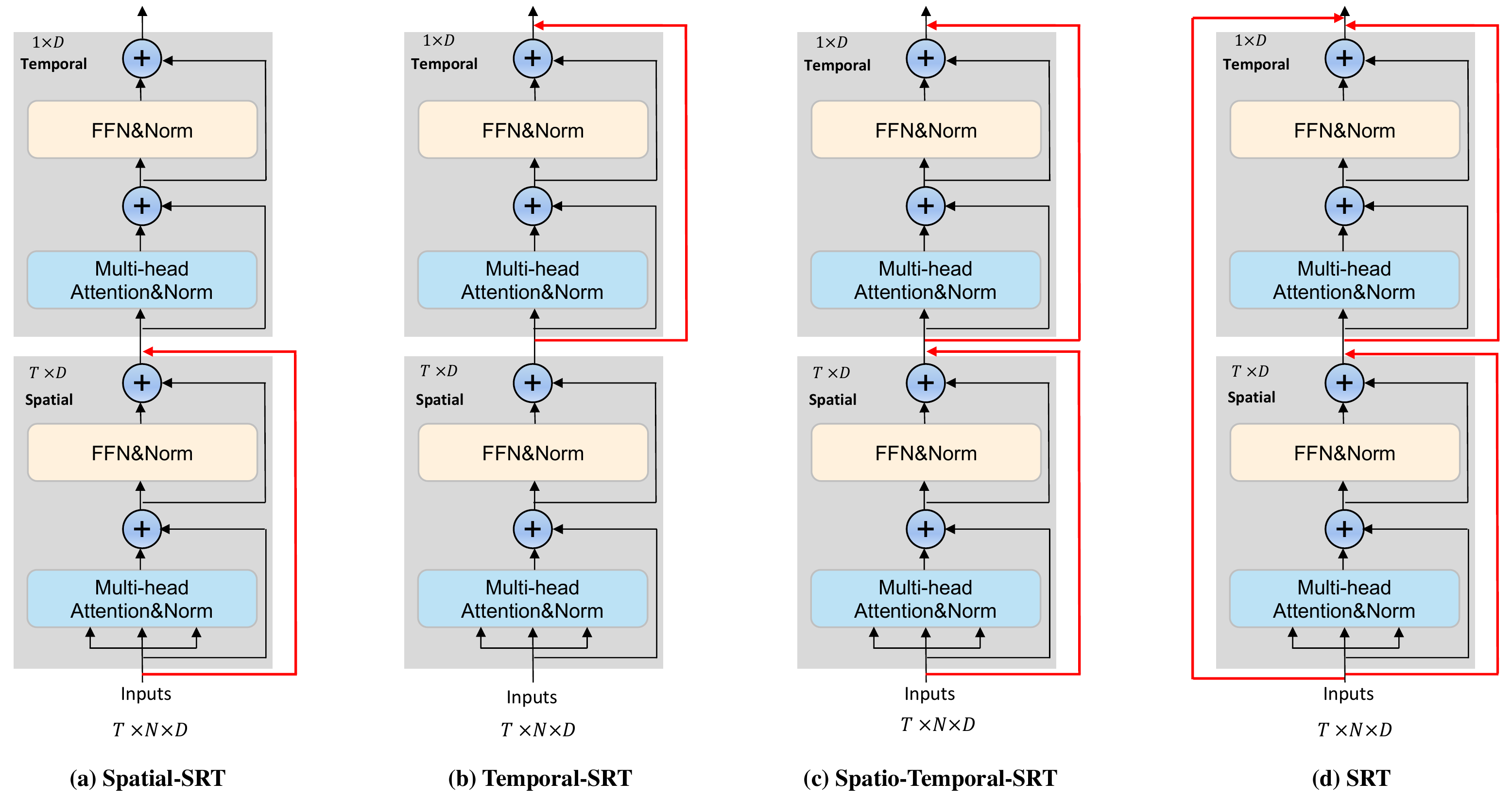}
   %\vspace{-0.1cm}
   \caption{Designs of SRT and its variants: (a) spatial residual (Spatial-SRT); (b) temporal residual (Temporal-SRT); (c) spatio-temporal residual (Spatio-Temporal-SRT); (d) our SRT.
}
   \label{fig:residual}
   %\vspace{-0.2cm}
\end{figure*}

\textbf{SRT for relation embedding learning.} We propose a spatio-temporal residual transformer architecture to learn local relation information in a video. In this spatio-temporal residual transformer, we have two data flows in which one flow operates across the frame and the other processes the object proposals inside each frame. Suppose that a set of object proposals $Y^{t}= \{ y^{t} \}^{N}_{n=1}$ are in frame $t$, where $y^{t}\in \mathbb{R}^{d_{r}}$ is the feature vector of the n-th proposal and $N$ is the top $N$ region proposals. We view each frame tensor $Y^{t}_{0}$ as a sequence of object proposal embeddings:
\begin{equation}
 Y^{t}_{0} = \left[ y^{t, 1}_{0}, y^{t, 2}_{0}, \cdot\cdot\cdot, y^{t, N}_{0} \right].
\end{equation}
For the object proposal embeddings, to capture spatial relations among visual objects, we utilize a transformer block to explore the interaction pattern in spatial (frame) between object proposals	. Then, a residual connection is used to aggregate spatial information and original local information:
\begin{equation}
 Y^{t'}_{l} = Y^{t}_{l-1} + MSA \left (LN \left(Y^{t}_{l-1} \right) \right),
\end{equation}
\begin{equation}
 Y^{t''}_{l} =  Y^{t'}_{l} + MLP \left(LN \left(Y^{t'}_{l} \right) \right),
\end{equation}
\begin{equation}
 Y^{t'''}_{l} =  Y^{t''}_{l} + Y^{t}_{l-1}.
\end{equation}
where $l = 1, 2, \cdot \cdot \cdot, L $ is index of the l-th layer, and L is the total number of layers. The updated features after multi-layer transformer block are forwarded to an average pooling layer, which calculates the mean of all the proposal features and leads to a $1 \times d^{r}$ dimensions representation. All frame tensors after transformation are 
\begin{equation}
Z_{0} = \left[Y^{1'''}_{L}, Y^{2'''}_{L}, \cdot\cdot\cdot, Y^{T'''}_{L} \right]. 
\end{equation}
This process builds the relationship among proposals by computing interactions between any two proposals. For the frame level, we create the object proposal embedding memories to store the sequence of frame-level representations $Z_{0}$. Similar to the object proposal level processing, we use a transformer block for transforming the frame embeddings. Then, a residual connection is used to aggregate temporal information with spatial information and spatio-temporal information with original local information, respectively. Our final relation embedding is defined as:
\begin{equation}
 Z^{'}_{l} = Z_{l-1} + MSA \left(LN \left (Z_{l-1}\right)\right),
\end{equation}
\begin{equation}
 Z^{''}_{l} = Z^{'}_{l} + MLP \left(LN \left(Z^{'}_{l}\right)\right),
\end{equation}
\begin{equation}
 Z^{'''}_{l} = Z^{''}_{l} + Z_{l-1},
\end{equation}
\begin{equation}
 F_{r} = Z^{'''}_{l} + Y^{t}_{l-1}.
\end{equation}
The temporal transformer block is used for modeling temporal relation among frame embeddings. Finally, we apply an attention-aware feature aggregation layer \cite{ging2020coot} to obtain the final relation embedding, denoted as $F_{r} \in \mathbb{R}^{d_{*}}$.

Next, we discuss several variants for SRT, as illustrated in Figure \ref{fig:residual}.
\textbf{Spatial-SRT} only utilizes Eq.(8) to aggregate spatial information and original local information by a residual connection (i.e., Figure (3a)). \textbf{Temporal-SRT} only adopts Eq.(12) to aggregate temporal information and spatial information by a residual connection (i.e., Figure (3b)). \textbf{Spatio-Temporal-SRT} only uses Eq.(8) and Eq.(12) to aggregate temporal information with spatial information and spatial information with original local information by a residual connection, respectively (i.e., Figure (3c)). Besides, we use \textbf{Non-SRT} as a base variant, and the module indicates that no residuals are added between the transformer blocks. We compare the above five variants of SRT on a standard benchmark in Section \ref{sec:as} and observe the SRT achieves the best performance. Moreover, we find that SRT introduces minor modifications of the residual connection but grants maximum benefits.

\subsection{Text Embedding Learning}\label{sec:tr}

For learning the contextual relations between the words in the video description sentence $s_{i}$, we adopt a BERT language representation model to encode the word sequence, and it applies the bidirectional training of transformer \cite{vaswani2017attention} to language modeling. It includes 12 layers of transformer blocks. Each block has 12 attention heads, and the hidden size is 768. Here, we take the hidden state of the per-token outputs of the last 2 layers to represent the information of the entire input sentence $F_{s} \in \mathbb{R}^{d_{t}}$. Finally, we transform each sentence representation $F_{s} \in \mathbb{R}^{d_{t}}$ into a text embedding feature $F_{t} \in \mathbb{R}^{d_{*}}$ by using a pointwise linear layer and an attention-aware feature aggregation layer \cite{ging2020coot}.

\subsection{Joint Embedding Learning}\label{sec:jel}

The purpose of joint embedding learning between video and textual features is to perform similarity comparisons. For a given video $V_{i}$, the proposed framework extracts two types of embedding features — video embeddings $F_{v}$ and relation embeddings $F_{r}$. We calculate the similarity between videos and sentences in both embedding spaces. Specifically, for a given sentence $T_{i}$, the similarity score with $V_{i}$ is obtained by summing the cosine similarities between its text embedding features $F_{t}$ and such two types of video embedding features, 
\begin{equation} \label{eqn:lamb}
S \left(V_{i}, T_{i} \right) = \lambda \cdot cosine \left(F_{r},  F_{t} \right) + \left(1-\lambda\right) \cdot cosine \left(F_{v}, F_{t}\right).
\end{equation}
where $0\leq \lambda \leq1$ is a hyper-parameter to balance the importance of two similarity scores. Based on the defined similarity score, we use a hinge-based triplet ranking loss to encourage the similarity score of matched video and sentence to be larger than those of mismatched ones:
\begin{equation} \label{eqn:delt}
\begin{aligned}
\mathcal{L}_{r}= \left[\delta-S \left(V_{i}, T_{i}\right)+S \left(V_{i}, T_{j}\right)\right]_{+}\\
 + \left[\delta-S\left(V_{i}, T_{i}\right)+S\left(V_{j}, T_{i}\right)\right]_{+},
\end{aligned}
\end{equation} where $0<\delta\leq1$ is the margin, the operator $[x]_{+} = max(x,0)$, and $S(\cdot,\cdot)$ is the similarity function.($V_{i},T_{i}$) represents the positive pair, while ($V_{i},T_{j}$) and ($V_{j},T_{i}$) represent the negative pairs available in the mini-batch.

% Ablation
\begin{table*}[t]
\centering
\begin{center}
\caption{Performance of introducing visual relations (VR) for cross-modal retrieval. The evaluations are done on 1k-A test set (Training-9k) \cite{gabeur2020multi} for MSR-VTT.}
\vspace{0.1cm}
\label{tab:abl}
\setlength{\tabcolsep}{5.5mm}{
\begin{tabular}{l|cccc|cccc}
\hline \hline
\multicolumn{1}{c|}{\multirow{2}{*}{Method}} & \multicolumn{4}{c|}{Text-to-Video} & \multicolumn{4}{c}{Video-to-Text} \\
\multicolumn{1}{c}{} & R@1 & R@5 & R@10 & MedR & R@1 & R@5 & R@10 & MedR  \\ 

\hline 

dataset split from \cite{gabeur2020multi} & & & &  & & & &\\

VG & 32.9  & 65.8 & 79.7  & 3 & 32.1 & 65.2  & 77.6 & 3 \\
$\rm VR_{st}$ & 29.8 & 64.4  & 77.2 & 3 & 29.2  & 63.6 &76.8  & 3 \\
$\rm VG+VR_m$ &32.7  & 66.0  &79.4  &3  & 32.9 & 67.0 & 79.6  & 3 \\
$\rm VG+VR_s$ &33.8  & 69.3  & 82.9 & 2 & 36.2 & 72.4  & 84.3 & 2 \\
$\rm VG+VR_t$ & 33.7 & 67.5  & 81.7 & 3 & 34.0 & 70.2  & 82.7 & 2 \\
%\hline 

BiC-Net    & \textbf{39.4} & \textbf{75.5} & \textbf{86.7} & \textbf{2} & \textbf{39.4} & \textbf{76.5} & \textbf{85.9} &  \textbf{2} \\
\hline \hline
\end{tabular}}
\end{center}
%\vspace{-0.5cm}
\end{table*}

% Ablation2
\begin{table*}[t]
\centering
\begin{center}
\caption{Performance of the variants for SRT for cross-modal retrieval. The evaluations are done on 1k-A test set (Training-9k) \cite{gabeur2020multi} for MSR-VTT.}
\vspace{0.1cm}
\label{tab:abl}
\setlength{\tabcolsep}{5.5mm}{
\begin{tabular}{l|cccc|cccc}
\hline \hline
\multicolumn{1}{c|}{\multirow{2}{*}{Method}} & \multicolumn{4}{c|}{Text-to-Video} & \multicolumn{4}{c}{Video-to-Text} \\
\multicolumn{1}{c}{} & R@1 & R@5 & R@10 & MedR & R@1 & R@5 & R@10 & MedR  \\ 

\hline 

dataset split from \cite{gabeur2020multi} & & & &  & & & &\\

Non-SRT & 36.2  & 73.9 & 84.4  & 2 & 38.2 & 74.3  & 87.5 & 2 \\
Spatial-SRT & 37.8 & 71.7  & 85.0 & 2 & 39.2  & 74.5 & 85.2  & 2 \\
Temporal-SRT & 37.8  & 71.9  & 85.2 & 2 & 40.1 & 73.9  & 85.8  & 2 \\
Spatio-Temporal-SRT & 38.2  & 73.2  & 85.8 & 2 & 39.3 & 74.2  & 85.5 & 2 \\
SRT (BiC-Net) & \textbf{39.4} & \textbf{75.5} & \textbf{86.7} & \textbf{2} & \textbf{39.4} & \textbf{76.5} & \textbf{85.9} &  \textbf{2} \\

\hline \hline
\end{tabular}}
\end{center}
%\vspace{-0.5cm}
\end{table*}

% result analysis

% MSR-VTT
\begin{table*}[t]
\centering
\begin{center}
\caption{Cross-modal retrieval comparison with state-of-the-art methods on MSR-VTT.}
\label{tab:msr}
\setlength{\tabcolsep}{4.9mm}{
\begin{tabular}{l|cccc|cccc}
\hline
\hline
\multicolumn{1}{c|}{\multirow{2}{*}{Method}} & \multicolumn{4}{c|}{Text-to-Video} & \multicolumn{4}{c}{Video-to-Text} \\
\multicolumn{1}{c|}{} & R@1 & R@5 & R@10 & MedR & R@1 & R@5 & R@10 & MedR  \\ 
\hline 
Full test set \cite{xu2016msr} & & & & & & & & \\
STG \cite{song2021spatial}  & 8.3 & 23.7 & 33.9 & 28  &\verb|-| &\verb|-| &\verb|-|  &\verb|-| \\
HGR \cite{chen2020fine} & 9.2 & 26.2 & 36.5 & 24 & 15.0 & 36.7 & 48.8 &  11   \\

DualEncoding \cite {dong2021dual} & 11.6 & 30.3 & 41.3 & 17  & \textbf{22.5} &  47.1 & 58.9 & 7  \\
T2VLAD \cite{wang2021t2vlad} & 12.7  & 34.8 & 47.1 & 12 & 20.7  & 48.9 & 62.1 & 6  \\

BiC-Net & \textbf{19.2} & \textbf{47.0} & \textbf{62.5} & \textbf{6} & 20.6 & \textbf{49.3} & \textbf{63.7} &  \textbf{6} \\

\hline

1k-B test set \cite{miech2018learning} & & & & & & & & \\

CE \cite{liu2019use}  & 18.2 & 46.0 & 60.7 & 7 & 18.0 & 46.0 & 60.3 & 6.5 \\
DualEncoding \cite {dong2021dual} & 23.0 & 50.6 & 62.5 & 5  & 25.1 &  52.1 & 64.6 & 5  \\

MMT \cite{gabeur2020multi}  & 20.3  & 49.1 & 63.9 & 6 & 21.1 & 49.4 & 63.2 & 6  \\

T2VLAD \cite{wang2021t2vlad} & 26.1  & 54.7 & 68.1 & 4 & 26.7  & 56.1 & 70.4 & 4  \\ 

BiC-Net  & \textbf{34.0} & \textbf{71.1} & \textbf{84.1} & \textbf{3} & \textbf{37.9} & \textbf{73.4} & \textbf{85.3} &  \textbf{2} \\

\hline 

1k-A test set (training-7k) \cite{miech2019howto100m} & & & & & & & & \\
%JSFusion \cite{Yu_2018_ECCV}   & 10.2 & 31.2 & 43.2 & 13 &\verb|-| &\verb|-| &\verb|-|  &\verb|-| \\
Miech et al. \cite{miech2019howto100m} & 12.1 & 35.0 & 48.0 & 12 &\verb|-| &\verb|-| &\verb|-|  &\verb|-| \\
STG \cite{song2021spatial}   & 15.5 & 39.2 & 50.4 & 10 &\verb|-| &\verb|-| &\verb|-|  &\verb|-| \\
TCE \cite{yang2020tree} & 17.1  & 39.9 & 53.7 & 9 &\verb|-| &\verb|-| &\verb|-|  &\verb|-|   \\
DualEncoding \cite {dong2021dual} & 21.6 & 49.5 & 62.3 & 6 & 27.8 & 48.7 &  58.7 & 6  \\
BiC-Net & \textbf{32.8} & \textbf{68.2} & \textbf{82.4} & \textbf{3} & \textbf{36.8} & \textbf{71.5} & \textbf{83.5} &  \textbf{2} \\

\hline 

1k-A test set (Training-9k) \cite{gabeur2020multi} & & & & & & & & \\

%CE \cite{liu2019use}  & 20.9  & 48.8 & 62.4 & 4 & 20.6 & 50.3 & 64.0 & 5.3 \\

MMT \cite{gabeur2020multi}  & 24.6  & 54.0 & 67.1 & 4 & 24.4 & 56.0 & 67.8 & 4  \\

SUPPORT-SET \cite{patrick2020support} & 27.4  & 56.3 & 67.7 & 3 & 26.6  & 55.1 & 67.5 & 3  \\

%T2VLAD \cite{wang2021t2vlad} & 29.5  & 59.0 & 70.1 & 4 & 31.8  & 60.0 & 71.1 & 3  \\
Frozen \cite{bain2021frozen} & 31.0               & 59.5  & 70.5  & 3 &\verb|-| &\verb|-| &\verb|-|  &\verb|-|   \\
    
%HCQ  \cite{wang2022hybrid}   &25.9               & 54.8   & 69.0  & 5.0  &\verb|-| &\verb|-| &\verb|-|  &\verb|-|   \\
       
CLIP4Clip \cite{luo2021clip4clip} & \textbf{44.5}  & 71.4 & 81.6 & 2 &\verb|-| &\verb|-| &\verb|-|  &\verb|-|   \\

BiC-Net  & 39.4 & \textbf{75.5} & \textbf{86.7} & \textbf{2} & \textbf{39.4} & \textbf{76.5} & \textbf{85.9} &  \textbf{2} \\

\hline \hline
\end{tabular}}
\end{center}
\vspace{-0.2cm}
\end{table*}

% MSVD
\begin{table*}[t]
\centering
\begin{center}
\caption{Cross-modal retrieval comparison with state-of-the-art methods on MSVD.}
\label{tab:msvd}
\setlength{\tabcolsep}{5.7mm}{
\begin{tabular}{l|cccc|cccc}
\hline
\hline
\multicolumn{1}{c|}{\multirow{2}{*}{Method}} & \multicolumn{4}{c|}{Text-to-Video} & \multicolumn{4}{c}{Video-to-Text} \\
\multicolumn{1}{c|}{} & R@1 & R@5 & R@10 & MedR & R@1 & R@5 & R@10 & MedR  \\ 

\hline 

Mithun et al. \cite {mithun2018learning} & 16.1 & 41.1 & 53.5 & 9 & 23.4 & 45.4 & 53.0 & 8 \\
CE \cite{liu2019use}  & 19.8  & 49.0 & 63.8 & 6 & \verb|-| &\verb|-| &\verb|-|  &\verb|-|   \\
ViSERN \cite{feng2020exploiting} & 18.1 & 48.4 & 61.3 & 6 & 24.3 & 46.2 & 59.5 & 7 \\
SUPPORT-SET \cite{patrick2020support} & 23.0  & 52.8 & 65.8 & 5 & \textbf{27.3}  & 50.7 & 60.8 & 5  \\

%\hline 
BiC-Net & \textbf{24.6}  & \textbf{57.0} & \textbf{70.3} & \textbf{4} &  24.2 & \textbf{58.7} & \textbf{70.1} & \textbf{4}  \\

\hline \hline
\end{tabular}}
\end{center}
\vspace{-0.3cm}
\end{table*}

% YouCook2
\begin{table*}[t]
\centering
\begin{center}
\caption{Cross-modal retrieval comparison with state-of-the-art methods on YouCook2. TS: trained from scratch on YouCook2.}
%PT: pretrained on the HowTo100M dataset. PF: pretrained on the HowTo100M dataset, then fine-tuned onYouCook2.
\label{tab:youcook2}
\setlength{\tabcolsep}{5.5mm}{
\begin{tabular}{l|cccc|cccc}
\hline
\hline
\multicolumn{1}{c|}{\multirow{2}{*}{Method}} & \multicolumn{4}{c|}{Text-to-Video} & \multicolumn{4}{c}{Video-to-Text} \\
\multicolumn{1}{c|}{} & R@1 & R@5 & R@10 & MedR & R@1 & R@5 & R@10 & MedR  \\ 

\hline 

HGLMM FV CCA \cite{miech2019howto100m}  & 4.6  & 14.3 & 21.6 & 75 & \verb|-| &\verb|-| &\verb|-|  &\verb|-|   \\
Miech et al. \cite{miech2019howto100m}  & 4.2 & 13.7 & 21.5 & 65  &\verb|-| &\verb|-| &\verb|-|  &\verb|-|   \\
COOT \cite{ging2020coot} & 5.9  & 16.7 & 24.8 & 49.7 & \verb|-| &\verb|-| &\verb|-|  &\verb|-|  \\
AME-Net \cite{han2022adversarial} & 7.6  & 21.5 & 32.8 & \textbf{28} &  7.9 & 22.5 & 32.2 & \textbf{28}  \\

BiC-Net (TS) &  \textbf{8.7}  &  \textbf{23.9} & \textbf{33.5} & 31 & \textbf{8.3}  & \textbf{23.6} & \textbf{32.6} & 31  \\

%ActBERT \cite{zhu2020actbert} & 9.6  & 26.7 & 38.0 & 19  & \verb|-| &\verb|-| &\verb|-|  &\verb|-| \\

%COOT (TF) \cite{ging2020coot}& 16.7 & 40.2  & 52.3 & 9 & \verb|-| &\verb|-| &\verb|-|  &\verb|-|  \\

%\hline 

%iC-Net (TF)  & \textbf{ }  & \textbf{ } & \textbf{ } & \textbf{} & \textbf{} & \textbf{} & \textbf{} & \textbf{ }  \\

\hline \hline
\end{tabular}}
\end{center}
\vspace{-0.3cm}
\end{table*}

\section{Experiments}\label{sec:exp}

\subsection{Experimental Setup}\label{sec:da}
1) Dataset: We evaluated the proposed BiC-Net model on three benchmarks: MSR-VTT, MSVD, and YouCook2. The MSR-VTT dataset \cite{xu2016msr} is the most widely-used dataset for text-video retrieval. It contains $10,000$ Youtube video clips with $20$ different text captions. Following the settings in \cite{xu2016msr, miech2018learning, Yu_2018_ECCV}, we adopt three kinds of evaluation settings. For the 1k-A test set \cite{Yu_2018_ECCV}, we using 7k train+val videos \cite{miech2019howto100m} and 9k train+val videos  for training \cite{gabeur2020multi} and report results. The MSVD dataset \cite{chen2011collecting} contains $1,970$ video clips from YouTube. Each video clip has around $40$ descriptions in multiple languages. We only adopt English annotations in experiments. Following prior work \cite {venugopalan2015sequence}, we separate the dataset into $1,200$ clips for training,, $100$ clips for validation, and $670$ clips for testing. The YouCook2 dataset \cite {zhou2018towards} contains $2,000$ cooking videos with $14,000$ video clips. It covers 89 types of recipes. Each video clip is described by a textual sentence. Referring to \cite{miech2019howto100m}, we evaluate the text-video clip retrieval task on the validation clips. 

2) Evaluation Metrics: We employ the widely used median retrieval rank (MedR) and recall rate at top $\mathcal{K}$ ($R@\mathcal{K}$) for assessing retrieval accuracy. MedR measures the median rank position among where true positives are returned. $R@\mathcal{K}$ measures the fraction of true positives being ranked at top $\mathcal{K}$ returned results. Therefore, lower MedR scores indicate higher performance; in contrast, higher $R@\mathcal{K}$ scores indicate better performance.

3) Implementation Details: We sample $26$ video frames from it with the same temporal duration between every two frames. In our experiments, the ILSVRC-2012-CLS \cite{russakovsky2015imagenet} pre-trained InceptionResNetV2 \cite{szegedy2017inception} is adopted to extract $1536$-D 2D features and the Kinetics \cite{kay2017kinetics} pre-trained I3D \cite{carreira2017quo} to extract $1024$-D 3D features. The number $N$ of regions within a frame is $36$, identical to \cite{anderson2018bottom}. The dimension d of region features extracted from ResNet-101 is $2048$-D. The dimensionality of video-embedding vectors $F_{v}$ and relation-embedding vectors $F_{r}$ are set as $1024$-D. For each sentence, we use pre-trained BERT to extract 1536-D word embedding and apply a pointwise linear layer and an attention-aware feature aggregation layer \cite{ging2020coot} to obtain $1024$-D text-embedding vectors.

We implement our proposed model using PyTorch\footnote{\url{http://www.pytorch.org}} and train it on 4 Tesla V100 GPUs. We train for 60 epochs using Adam optimizer \cite{kingma2015adam} with a mini-batch size of $64$. On the MSR-VTT, MSVD, and YouCook2, the learning rates are set to 0.0002, 0.0004, and 0.0004,  respectively. As for the layer number L of transformer block, we set it to $4$, $2$  and $4$ on the MSR-VTT, MSVD,  and YouCook2 datasets, respectively. In addition, the trade-off parameter $\lambda$ in Eq. (\ref{eqn:lamb}), the margin $\delta$ in Eq. (\ref{eqn:delt}) are set to $0.5$ and $0.2$, respectively.

\subsection{Ablation Studies}\label{sec:as}

1) \textbf{Experiments with spatio-temporal Relation.} We experimented with variants of our model to verify the effectiveness of introducing spatio-temporal relation for text-video retrieval:

\begin{itemize}

\item $\mathbf{VG}$. We only utilize the pre-trained 2D and 3D CNNs to extract the global features of the whole video as video embedding learning.

\item $\mathbf{VG+VR_m}$. We apply the average-pooling features of all regions without using the features extracted by spatio-temporal residual transformer as relation embedding learning.

\item $\mathbf{VG+VR_s}$. We only utilize regional spatial relation features as relation embedding learning and global features as video embedding learning.

\item $\mathbf{VG+VR_t}$. We only utilize regional temporal relation features as relation embedding learning and global features as video embedding learning.

\item $\mathbf{VR_{st}}$. We only utilize regional spatio-temporal relation features as relation embedding learning.

\end{itemize}

We explore these model variants on the MSR-VTT, as shown in Table \ref{tab:abl}. We omit the results on MSVD and YouCook2 because of space limitations, but they show similar trends to MSR-VTT. From the results, we have the following observations. First, as expected, on both text-to-video and video-to-text, our BiC-Net, $\mathbf{VG+VR_s}$, and $\mathbf{VG+VR_t}$ significantly outperforms $\mathbf{VG}$ alone. The result verifies the significance of introducing spatio-temporal relation representation. Second, compared with $\mathbf{VG+VR_m}$, the performance of our BiC-Net verifies that the spatio-temporal residual transformer can capture the fine-grained local relational features. Third, compared with two variants of models (i.e., $\mathbf{VG}$ and $\mathbf{VR_{st}}$) that only use either global visual features or local relational features, our model considers both global and local relational features to achieve the best performance. This verifies the effectiveness of aligning the global visual and local relational features with text features on two embedding spaces. Notably, global visual features and local relational features are highly complementary, and their combination leads to an improvement far beyond the performance of the global visual features alone. Moreover, compared with $\mathbf{VG+VR_s}$ and $\mathbf{VG+VR_t}$, our BiC-Net achieves substantially better performance, which reveals the complementary of the spatial and temporal relation features.

2) \textbf{Evaluation of of SRT.} We test the effectiveness of our proposed  SRT and its variants on relation embedding learning. As shown in Table IV, our SRT and its variants achieve better performance than Non-SRT, which indicates the effectiveness of spatio-temporal relation modeling by adding residual blocks. The difference between Spatial-SRT and Temporal-SRT is that a residual block is added at different positions.  We can see that Temporal-SRT significantly surpasses Spatial-SRT, which indicates the importance of temporal relation modeling. Spatio-Temporal-SRT adds a residual block based on Spatial-SRT/Temporal-SRT, which achieves better performance than Spatial-SRT/Temporal-SRT by aggregating temporal information with spatial information. In the end, compared to the other variants, we observed that our proposed SRT achieves the best performance when three residual blocks are added, indicating that simultaneously adding residual blocks in our model performs better than adding only one of them. To sum up, the contribution of each component enables our SRT to learn higher-level spatio-temporal relation information.

\subsection{Comparison with State-of-the-art Methods}\label{sec:cs}
To demonstrate the effectiveness of the BiC-Net solution, we compared it to several state-of-the-art baselines: (1) RNN-based methods: DualEncoding \cite{dong2021dual}, TCE \cite{yang2020tree}, (2) Multimodal Fusion methods: Mithun et al. \cite {mithun2018learning}, CE \cite{liu2019use}, MMT \cite{gabeur2020multi}, (3) GCN-based methods: ViSERN, \cite{feng2020exploiting}, STG \cite{song2021spatial} and AME-Net \cite{han2022adversarial}, (4) Transformer-based methods: COOT \cite{ging2020coot}, CLIP4Clip \cite{luo2021clip4clip} and Frozen \cite{bain2021frozen}, (5) other methods: HGLMM FV CCA \cite{miech2019howto100m}, Miech et al. \cite{miech2019howto100m},  SUPPORT-SET \cite{patrick2020support}, T2VLAD \cite{wang2021t2vlad}. 

%Note that for fair comparisons, we directly cited the results from their original papers without pre-training on HowTo100M \cite{miech2019howto100m}. 

1) Experiments on MSR-VTT: The experimental results are presented in Table \ref{tab:msr}. We can observe that for all data partitions, our proposed method consistently outperforms all, compared to traditional RNN-based methods and multimodal Fusion methods in all evaluation metrics by a large margin, including CE \cite{liu2019use}, MMT \cite{gabeur2020multi}, and T2VLAD \cite{wang2021t2vlad}, which use expert features (e.g., object, motion, face, scene, sound, and speech). Moreover, our BiC-Net significantly outperforms recent spatio-temporal relation-based method (STG \cite{song2021spatial}) in all evaluation metrics, especially, it boosts the text-video retrieval quality by a margin of 28.6\% in R@10 on full test set. This condition reveals the effectiveness of our BiC for modeling video global and relational information. The obvious performances are shown on full test set and 1k-A test set (training-7k).

We also compare with some typical transformer-based methods, such as CLIP4Clip \cite{luo2021clip4clip} and Frozen \cite{bain2021frozen}. CLIP4Clip adopts the language-vision transformer model of CLIP \cite{radford2021learning} pre-trained on a large-scale text-image dataset as a backbone. Frozen uses a transformer-based video model \cite{bertasius2021space} as a backbone. In contrast, we design a new transformer-based backbone to model spatio-temporal relations and global temporal information. Our BiC outperforms most of the compared methods on 1k-A test set (Training-9k), e.g., BiC 86.7\% \emph{vs} CLIP4Clip 81.6\% w.r.t. text-to-video R@10. This indicates that learning cross-modal complementarity in a cooperative and complementary manner takes effect.

2) Experiments on MSVD: Table \ref{tab:msvd} summarizes the performance comparison results. We also observe that our proposed BiC outperforms recent state-of-the-art methods in terms of most indicators. Note that among all these methods, ViSERN \cite{feng2020exploiting} uses only local video features to compute the similarity between the video and text. Analogously, we also observe that BiC-Net outperforms the local feature-based method ViSERN \cite{feng2020exploiting} by a great margin. This reveals that jointly modeling the global and local video representation plays a significant role in text-video retrieval, contributing to more powerful representation. To ensure a fair comparison, we compare the previous SOTA method, SUPPORT-SET \cite{patrick2020support} without pre-training on HowTo100M \cite{miech2019howto100m}. Under the full fair comparison, our BiC outperforms the previous best method SUPPORT-SET by 9.3\% on video-text retrieval R@10. Notably, on MSVD, the performance of our model is not as outstanding. The reason for small gains is that the transformer has the property of lacking structural bias making it prone to overfitting for small-scale data.

3) Experiments on YouCook2: As shown in Table \ref{tab:youcook2}, our method achieves the best performance, which is 8.7\% absolute gains in the evaluation metric of text-video retrieval R@10 better than COOT~\cite{ging2020coot}. In Miech et al. \cite{miech2019howto100m} and COOT~\cite{ging2020coot}, the global video features are used for video representation, whose performances are worse than most methods. AME-Net~\cite{han2022adversarial} adopts global features and handcrafted graph-based relation features. AME-Net achieves better performances than Miech et al. \cite{miech2019howto100m}, while their performances are worse than ours, which indicates that the global and local information learned by our method can be mutually promoted in a complementary manner. This observation indicates that in addition to the global video features, local relation features are also important for video representation. 

%In addition, training our approach with HowTo100M \cite{miech2019howto100m} pretrained features yields considerable performance improvement, compared to the pretraining approach (COOT (TF)). 

\begin{figure}[t]
   \center
   \includegraphics[width=0.46\textwidth]{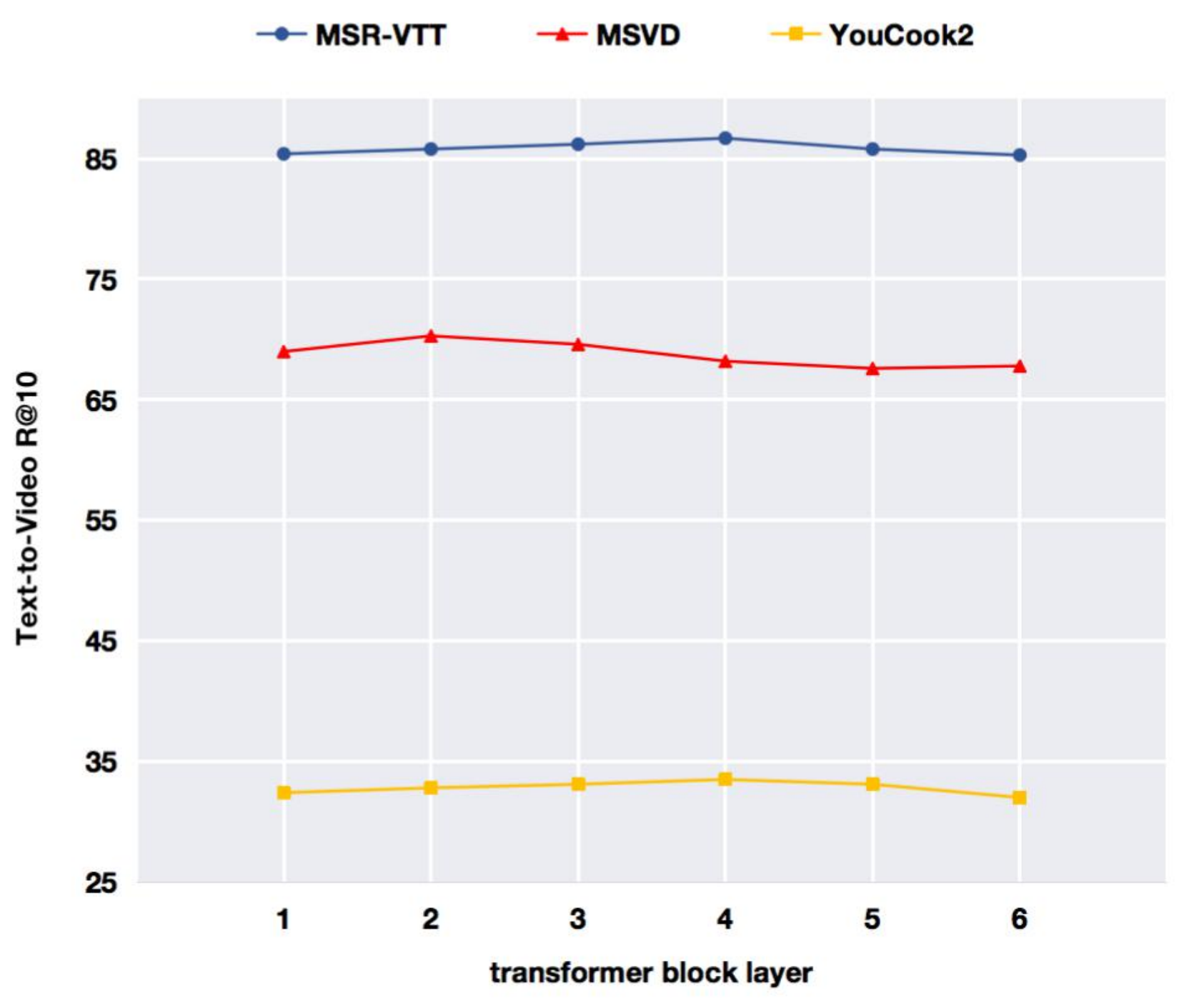}
   %\vspace{-0.1cm}
   \caption{Performance of text-to-video retrieval with different layers number of transformer blocks. }
   \label{fig:layers}
   \vspace{-0.25cm}
\end{figure}

\begin{figure}[t]
   \center
   \includegraphics[width=0.46\textwidth]{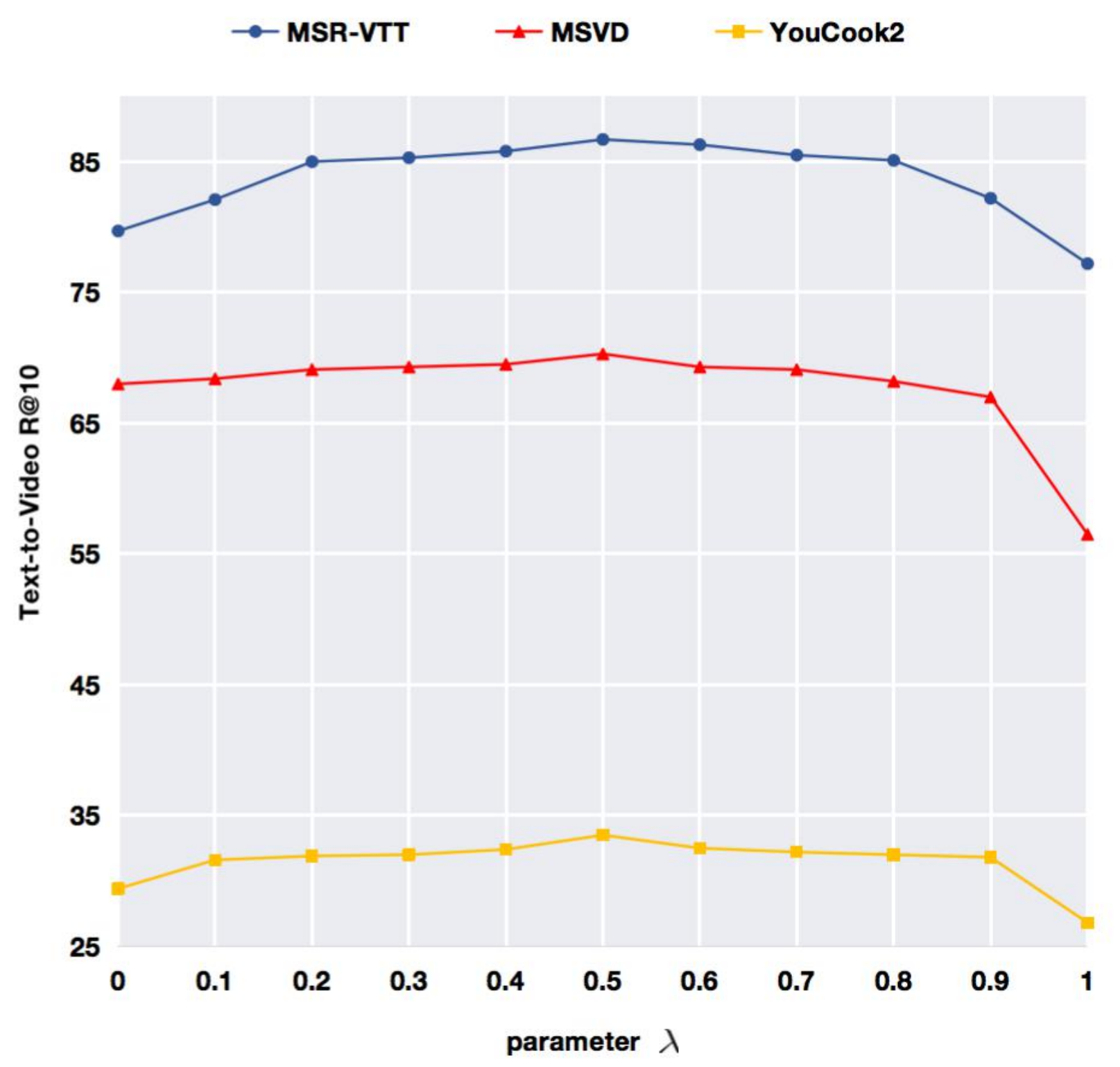}
   %\vspace{-0.1cm}
   \caption{Evaluation of different weight combination of the global and relation similarities.}
   \label{fig:parameter-off}
   \vspace{-0.25cm}
\end{figure}

\subsection{Parametric Sensitivity Analysis}

We carry out experiments to explore how the layer number of transformer block L and the trade-off parameter $\lambda$ affect the retrieval performance. Notably, we omit the video-text retrieval results on three datasets due to space limitations, which show similar trends to text-video retrieval. First, we analyze the influence of the hyperparameter, that is, the layer number of transformer block on the MSR-VTT 1k-A test set \cite{gabeur2020multi}, MSVD, and YouCook2 datasets. Figure \ref{fig:layers} presents the results across the layer number of transformer block on the two datasets by R@10; note that R@1 and R@5 present the same trend, in which performances increase until certain numbers ($4$, $2$, and $4$ for MSR-VTT, MSVD and YouCook2 datasets, respectively) and then become stable. This result is due to the model's capability of capturing the spatio-temporal relations of the deepest layer numbers. 

Moreover, the influence of the hyperparameter $\lambda$ in Eq. (11) is revealed in Figure \ref{fig:parameter-off}. We assign different trade-off parameter $\lambda$ to the two scores (i.e., $\mathbf{VR_{st}}$ and $\mathbf{VG}$) to observe their influence on the matching performance on the three datasets. By analyzing the results shown in Figure \ref{fig:parameter-off}, we have the following observations: 1) The leftmost part of Figure \ref{fig:parameter-off} shows the results when $\mathbf{VR_{st}}$ accounts for 0, that is, when the proportion of $\mathbf{VG}$ is 1, which means that we remove the visual relations module from our model (i.e., $\mathbf{VG}$). We can observe that when the spatio-temporal relations module is removed, the retrieval performance is reduced by a large margin over the three datasets. This condition shows the positive effect of comprehensively introducing spatio-temporal relations for text-video retrieval. 2) Increasing the proportion of $\mathbf{VR_{st}}$ substantially boosts the performance of model. Our model performs best performance on the three datasets when $\lambda$ = $0.5$. Therefore, we argue that $\mathbf{VR_{st}}$ and $\mathbf{VG}$ occupy the same contribution to the overall similarity. We conclude that the two similarities work together to obtain the best retrieval performance in a cooperative manner.

\begin{table}
\begin{center}
\caption{Comparison with different models in terms of model size and computation overhead at the inference stage.}
\setlength{\tabcolsep}{5.5mm}{
\label{tab:complexity}
\begin{tabular}{l|c|c}
\hline  \hline
Model &  Parameters (M) & FLOPs (G)  \\ 
\hline 

MMT \cite{gabeur2020multi}   & 133.4 & 12.64  \\
DualEncoding \cite {dong2021dual}    & 95.9  & \textbf{3.64}  \\

BiC-Net  & \textbf{31.48} & 10.33  \\

\hline \hline

\end{tabular}}
\end{center}
\end{table}

\subsection{Model Complexity}

We compare our method with open-source methods in terms of model size and computation overhead at the inference stage. As shown in Figure \ref{fig:layers}, since the performance of using one layer of transformer block outperforms MMT by a large margin, we only calculate the model size and computational overhead of using one layer of transformer block. Analogously, we also observe that our BiC-Net with one layer of transformer block outperforms DualEncoding by a great margin on the MSR-VTT 1k-A test set \cite{miech2019howto100m}. Notably, we omit the text-video retrieval results of VSR with a layer of transformer block on the MSR-VTT 1k-A test set \cite{miech2019howto100m} due to space limitations, which show similar trends to the MSR-VTT 1k-A test set \cite{gabeur2020multi}. In addition, we conclude that for each additional layer of transformer block, the computational cost will increase by $8.29$ GFLOPs and the parameters will increase by $25.19$M. Following \cite {dong2021dual}, we measure the number of FLOPs required for a text-video pair. As shown in Table \ref{tab:complexity} and Figure \ref{fig:layers}, we have two main observations: 1) our BiC-Net with one layer of transformer block achieves 85.6\% text-to-video R@10 accuracy on the MSR-VTT 1k-A test set \cite{gabeur2020multi}, which is 18.5\% higher than MMT, with fewer parameters and lower computational cost. 2) our BiC-Net with one layer of transformer block is smaller and slightly slower than DualEncoding.

\begin{figure*}[t]

    \centering
    \subfloat[]{\includegraphics[width=0.5\textwidth]{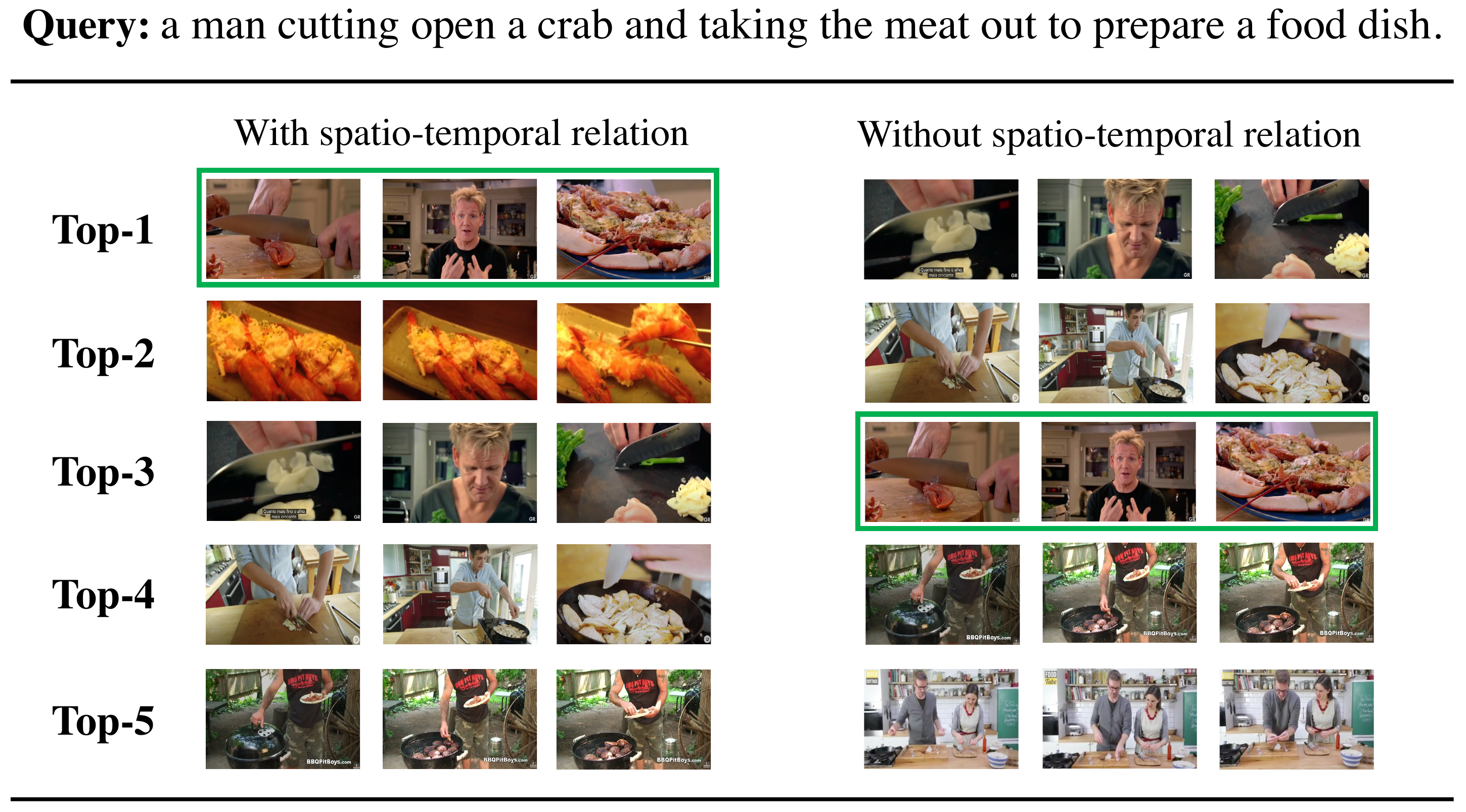}
    \label{fig_alpha1}}
    %\hfil
    \subfloat[]{\includegraphics[width=0.5\textwidth]{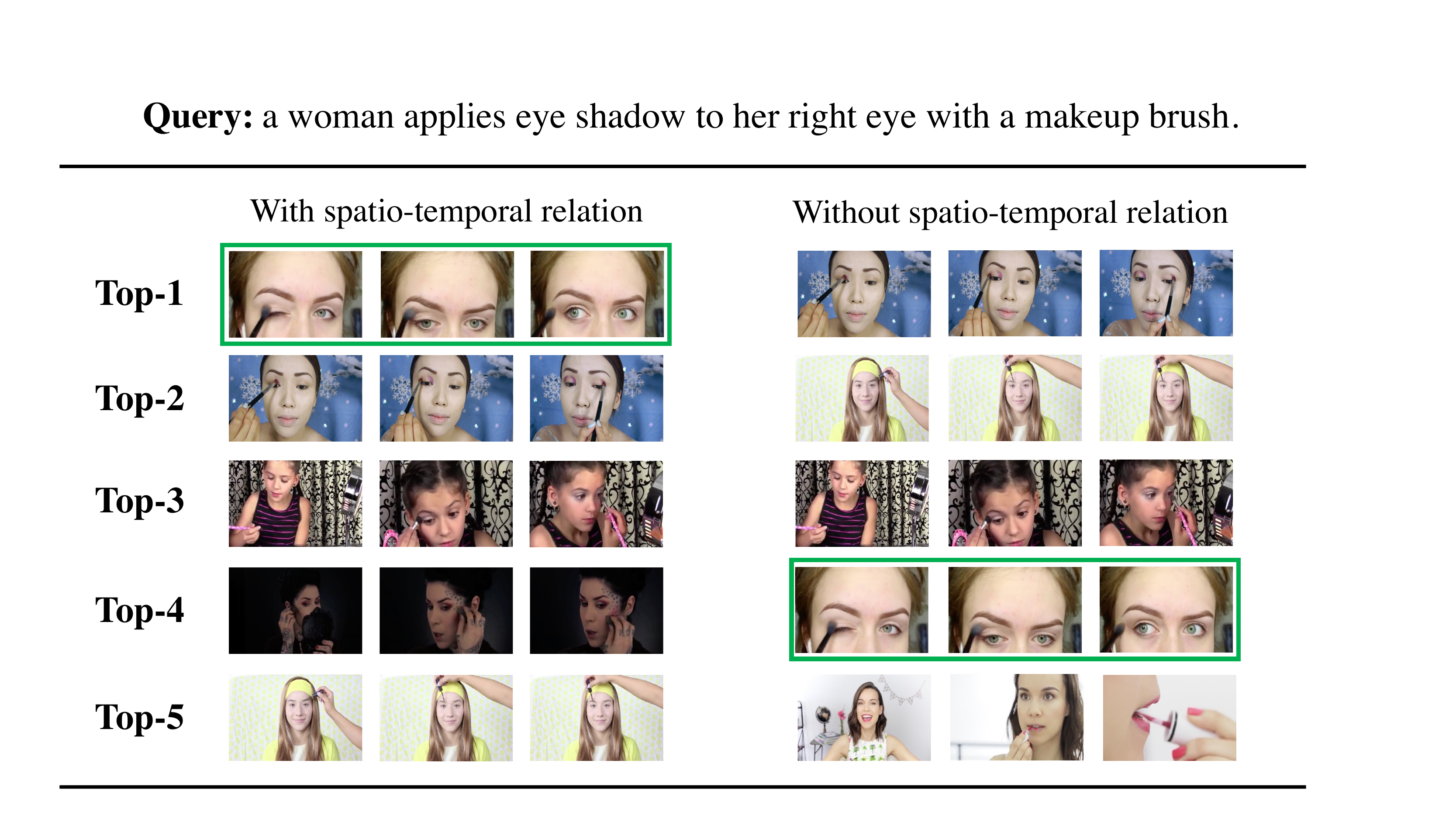}
    \label{fig_alpha2}}
    %\vspace{-0.3cm}
    \caption{Qualitative examples of the text-video tasks: In (a), (b), we show retrieval ranks of BiC-Net and the variant VG on MSR-VTT dataset test set \cite{miech2019howto100m}. Given a textual description as a query, we retrieve the most relevant video ranked from top to bottom. True positives are bounded in green boxes.}
    \label{fig_alpga}
    \vspace{-0.2cm}
\end{figure*}

\subsection{Qualitative Results}

Figure \ref{fig_alpga} shows two examples of text-to-video retrieval results between the model with and without visual spatio-temporal relations. We specifically choose two text-video retrieval examples that include complex spatio-temporal relationships. Figure \ref{fig_alpga} $(a)$ shows multiple sub-actions retrieval example; the sentence describes two objects (``man'' and ``crab''), and two actions (``cutting open a crab'' and ``taking the meat out'') in a short-term segment, which requires accurate spatio-temporal grounding. Comparing BiC-Net to its variant \textbf{VG}, our model successfully retrieves the correct video, which contains all spatio-temporal relationships and entities described in the sentence. The second video only contains ``taking the meat out from carb'' actions. The third and fourth videos only involve a ``cutting'' action and similar objects (e.g., ``man'' and ``knife''). The fifth video also only contains an action (``taking the meat''). In the left example, the \textbf{VG} model also retrieves similar scenes (e.g., similar man and cutting action) in the video. However, we observe that videos involving related elements are only ranked as the true positive in the top-3 positions. The performance of \textbf{VG} indicates that removing the fine-grained spatio-temporal relationships hurts the expressiveness of the video representation and further degrades the retrieval performance. Another example is showed in Figure \ref{fig_alpga} $(b)$, which requires fine-grained spatio-temporal relation grounding. The positive example contains a scenario involving two objects (``woman'', and ``makeup brush'') and a fine-grained action (``applies eye shadow''). Comparing these results, we observe that the variant \textbf{VG} retrieves a list of similar action videos, which cannot capture the fine-grained action (``applies eye shadow to her right eye'') in the video. Our model not only identifies the relevant objects ``woman'' and ``makeup brush'' but also captures the fine-grained relations between them. Again, it verifies the effectiveness of introducing spatio-temporal relation features to distinguish videos with the same visual components but with different relations. 

\begin{figure*}[t]
   \center
   \includegraphics[width=0.98\textwidth]{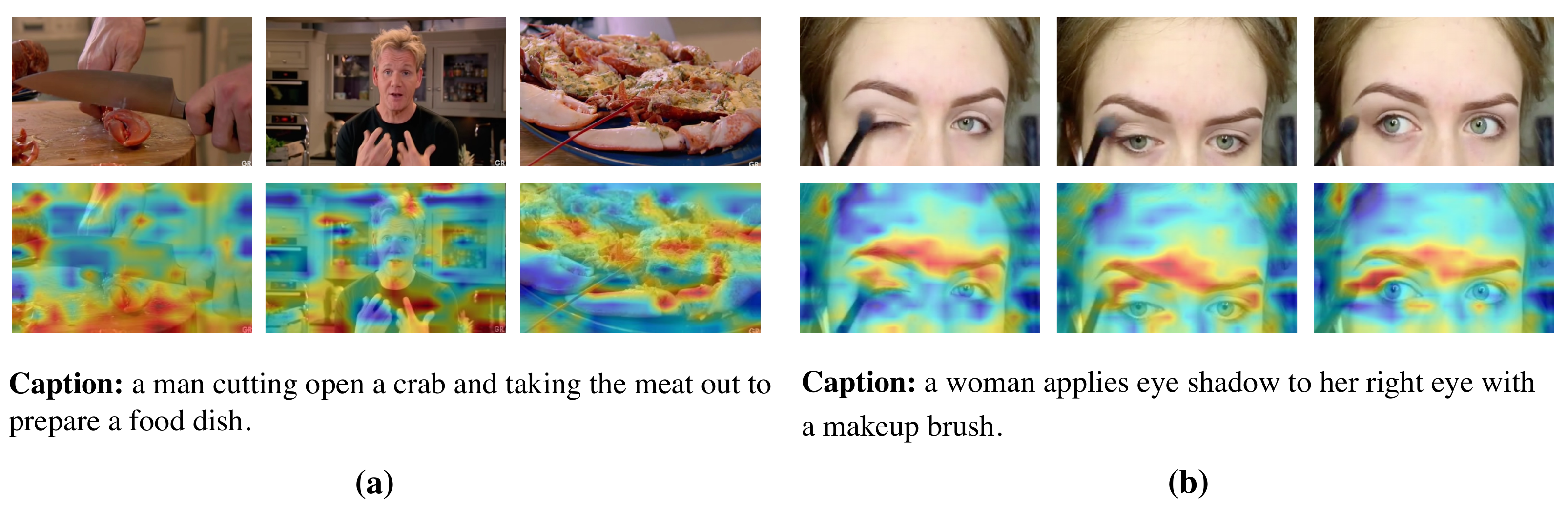}
   %\vspace{-0.1cm}
  \caption{Visualization of attention map on sample clips from the MSR-VTT. The top row presents original frames, and the bottom presents corresponding attention maps.}
   \label{fig:attention_map}
   \vspace{-0.25cm}
\end{figure*}

\subsection{Visualization Results}

To intuitively observe the effectiveness of introducing spatio-temporal relations, we visualize the attention map to infer the value of the spatio-temporal relation features. We select 2 videos, including two positive example in text-to-video retrieval from MSR-VTT. In Figure \ref{fig:attention_map}, we show the original frames and attention maps. As can be seen, our BiC-Net learns to value core parts with intense semantic relations such as ``man + crab'' in ``cutting open a crab and taking the meat out'', ``woman + makeup brush'' in ``applies eye shadow to her right eye”.  Furthermore, we find that the salient regions (e.g., man, crab, woman's right eye) are highlighted  separately in Figure \ref{fig:attention_map}. This also verifies that our model can learn fine-grained relational information with the corresponding text sentences.

\section{Conclusions}\label{sec:con}

This work contributes to a novel modeling method for cross-modal text-video retrieval. We claim that video representation should learn not only from global features but also from local spatio-temporal relationships. To fulfill this target, we design the Bi-Branch Complementary Network (BiC-Net) to capture local relational and global visual information for modeling comprehensively. Extensive experimental results on three benchmarks have demonstrated the effectiveness and superiority of our proposed method. Besides, we still face an inherent computational burden of attention in processing long-length video with more complex local relations. Therefore, we leave computational optimization of the multi-layer spatio-temporal transformer as future works.

\ifCLASSOPTIONcaptionsoff
  \newpage
\fi



\bibliographystyle{IEEEtran}


\bibliography{IEEEfull,ref}

\begin{thebibliography}{10}
\providecommand{\url}[1]{#1}
\csname url@samestyle\endcsname
\providecommand{\newblock}{\relax}
\providecommand{\bibinfo}[2]{#2}
\providecommand{\BIBentrySTDinterwordspacing}{\spaceskip=0pt\relax}
\providecommand{\BIBentryALTinterwordstretchfactor}{4}
\providecommand{\BIBentryALTinterwordspacing}{\spaceskip=\fontdimen2\font plus
\BIBentryALTinterwordstretchfactor\fontdimen3\font minus
  \fontdimen4\font\relax}
\providecommand{\BIBforeignlanguage}[2]{{%
\expandafter\ifx\csname l@#1\endcsname\relax
\typeout{** WARNING: IEEEtran.bst: No hyphenation pattern has been}%
\typeout{** loaded for the language `#1'. Using the pattern for}%
\typeout{** the default language instead.}%
\else
\language=\csname l@#1\endcsname
\fi
#2}}
\providecommand{\BIBdecl}{\relax}
\BIBdecl

\bibitem{song2021spatial}
X.~Song, J.~Chen, Z.~Wu, and Y.-G. Jiang, ``Spatial-temporal graphs for
  cross-modal text2video retrieval,'' \emph{IEEE Transactions on Multimedia},
  2021.

\bibitem{miech2019howto100m}
A.~Miech, D.~Zhukov, J.~Alayrac, M.~Tapaswi, I.~Laptev, and J.~Sivic,
  ``Howto100m: Learning a text-video embedding by watching hundred million
  narrated video clips,'' in \emph{Proceedings of the IEEE/CVF International
  Conference on Computer Vision, {(ICCV)}}, 2019, pp. 2630--2640.

\bibitem{feng2020exploiting}
Z.~Feng, Z.~Zeng, C.~Guo, and Z.~Li, ``Exploiting visual semantic reasoning for
  video-text retrieval,'' in \emph{International Joint Conference on Artificial
  Intelligence, {(IJCAI)}}, 2020, pp. 1005--1011.

\bibitem{chen2020fine}
S.~Chen, Y.~Zhao, Q.~Jin, and Q.~Wu, ``Fine-grained video-text retrieval with
  hierarchical graph reasoning,'' in \emph{Proceedings of the IEEE/CVF
  Conference on Computer Vision and Pattern Recognition, {(CVPR)}}, 2020, pp.
  10\,635--10\,644.

\bibitem{wu2021hanet}
P.~Wu, X.~He, M.~Tang, Y.~Lv, and J.~Liu, ``Hanet: Hierarchical alignment
  networks for video-text retrieval,'' in \emph{Proceedings of the ACM
  international conference on Multimedia}, 2021.

\bibitem{ren2015faster}
S.~Ren, K.~He, R.~Girshick, and J.~Sun, ``Faster r-cnn: Towards real-time
  object detection with region proposal networks,'' \emph{Advances in neural
  information processing systems}, vol.~28, 2015.

\bibitem{yang2020tree}
X.~Yang, J.~Dong, Y.~Cao, X.~Wang, M.~Wang, and T.-S. Chua, ``Tree-augmented
  cross-modal encoding for complex-query video retrieval,'' in
  \emph{Proceedings of the 43rd international ACM SIGIR conference on research
  and development in information retrieval}, 2020, pp. 1339--1348.

\bibitem{dong2021dual}
J.~Dong, X.~Li, C.~Xu, X.~Yang, G.~Yang, X.~Wang, and M.~Wang, ``Dual encoding
  for video retrieval by text,'' \emph{IEEE Transactions on Pattern Analysis
  and Machine Intelligence}, 2021.

\bibitem{dong2018predicting}
J.~Dong, X.~Li, and C.~G. Snoek, ``Predicting visual features from text for
  image and video caption retrieval,'' \emph{IEEE Transactions on Multimedia},
  vol.~20, no.~12, pp. 3377--3388, 2018.

\bibitem{liu2019use}
Y.~Liu, S.~Albanie, A.~Nagrani, and A.~Zisserman, ``Use what you have: Video
  retrieval using representations from collaborative experts,'' in
  \emph{British Machine Vision Conference}, 2019, p. 279.

\bibitem{li2020sea}
X.~Li, F.~Zhou, C.~Xu, J.~Ji, and G.~Yang, ``Sea: Sentence encoder assembly for
  video retrieval by textual queries,'' \emph{IEEE Transactions on Multimedia},
  vol.~23, pp. 4351--4362, 2020.

\bibitem{miech2018learning}
A.~Miech, I.~Laptev, and J.~Sivic, ``Learning a text-video embedding from
  incomplete and heterogeneous data,'' \emph{arXiv preprint arXiv:1804.02516},
  2018.

\bibitem{dong2019dual}
J.~Dong, X.~Li, C.~Xu, S.~Ji, Y.~He, G.~Yang, and X.~Wang, ``Dual encoding for
  zero-example video retrieval,'' in \emph{Proceedings of the IEEE/CVF
  Conference on Computer Vision and Pattern Recognition, {(CVPR)}}, 2019, pp.
  9346--9355.

\bibitem{wray2019fine}
M.~Wray, D.~Larlus, G.~Csurka, and D.~Damen, ``Fine-grained action retrieval
  through multiple parts-of-speech embeddings,'' in \emph{Proceedings of the
  IEEE/CVF International Conference on Computer Vision, {(ICCV)}}, 2019, pp.
  450--459.

\bibitem{han2021fine}
N.~Han, J.~Chen, G.~Xiao, H.~Zhang, Y.~Zeng, and H.~Chen, ``Fine-grained
  cross-modal alignment network for text-video retrieval,'' in
  \emph{Proceedings of the ACM international conference on Multimedia}, 2021,
  pp. 3826--3834.

\bibitem{han2022adversarial}
N.~Han, J.~Chen, H.~Zhang, H.~Wang, and H.~Chen, ``Adversarial multi-grained
  embedding network for cross-modal text-video retrieval,'' \emph{ACM
  Transactions on Multimedia Computing, Communications, and Applications},
  vol.~18, no.~2, pp. 1--23, 2022.

\bibitem{mithun2018learning}
N.~C. Mithun, J.~Li, F.~Metze, and A.~K. Roy-Chowdhury, ``Learning joint
  embedding with multimodal cues for cross-modal video-text retrieval,'' in
  \emph{Proceedings of the 2018 ACM on International Conference on Multimedia
  Retrieval}, 2018, pp. 19--27.

\bibitem{gabeur2020multi}
V.~Gabeur, C.~Sun, K.~Alahari, and C.~Schmid, ``Multi-modal transformer for
  video retrieval,'' in \emph{Proceedings of the European Conference on
  Computer Vision, (ECCV)}, vol.~5, 2020.

\bibitem{wang2021t2vlad}
X.~Wang, L.~Zhu, and Y.~Yang, ``T2vlad: global-local sequence alignment for
  text-video retrieval,'' in \emph{Proceedings of the IEEE Conference on
  Computer Vision and Pattern Recognition, {(CVPR)}}, 2021, pp. 5079--5088.

\bibitem{miech2020end}
A.~Miech, J.-B. Alayrac, L.~Smaira, I.~Laptev, J.~Sivic, and A.~Zisserman,
  ``End-to-end learning of visual representations from uncurated instructional
  videos,'' in \emph{Proceedings of the IEEE/CVF Conference on Computer Vision
  and Pattern Recognition, {(CVPR)}}, 2020, pp. 9876--9886.

\bibitem{rouditchenko2020avlnet}
A.~Rouditchenko, A.~Boggust, D.~Harwath, D.~Joshi, S.~Thomas, K.~Audhkhasi,
  R.~Feris, B.~Kingsbury, M.~Picheny, A.~Torralba \emph{et~al.}, ``Avlnet:
  Learning audio-visual language representations from instructional videos,''
  \emph{arXiv preprint arXiv:2006.09199}, 2020.

\bibitem{patrick2020support}
M.~Patrick, P.-Y. Huang, Y.~Asano, F.~Metze, A.~G. Hauptmann, J.~F. Henriques,
  and A.~Vedaldi, ``Support-set bottlenecks for video-text representation
  learning,'' in \emph{International Conference on Learning Representations,
  {(ICLR)}}, 2021.

\bibitem{luo2021clip4clip}
H.~Luo, L.~Ji, M.~Zhong, Y.~Chen, W.~Lei, N.~Duan, and T.~Li, ``Clip4clip: An
  empirical study of clip for end to end video clip retrieval,'' \emph{arXiv
  preprint arXiv:2104.08860}, 2021.

\bibitem{lei2021less}
J.~Lei, L.~Li, L.~Zhou, Z.~Gan, T.~L. Berg, M.~Bansal, and J.~Liu, ``Less is
  more: Clipbert for video-and-language learning via sparse sampling,'' in
  \emph{Proceedings of the IEEE/CVF Conference on Computer Vision and Pattern
  Recognition, {(CVPR)}}, 2021, pp. 7331--7341.

\bibitem{fang2021clip2video}
H.~Fang, P.~Xiong, L.~Xu, and Y.~Chen, ``Clip2video: Mastering video-text
  retrieval via image clip,'' \emph{arXiv preprint arXiv:2106.11097}, 2021.

\bibitem{liu2021hit}
S.~Liu, H.~Fan, S.~Qian, Y.~Chen, W.~Ding, and Z.~Wang, ``Hit: Hierarchical
  transformer with momentum contrast for video-text retrieval,'' in
  \emph{Proceedings of the IEEE/CVF International Conference on Computer
  Vision, {(CVPR)}}, 2021, pp. 11\,915--11\,925.

\bibitem{bain2021frozen}
M.~Bain, A.~Nagrani, G.~Varol, and A.~Zisserman, ``Frozen in time: A joint
  video and image encoder for end-to-end retrieval,'' in \emph{Proceedings of
  the IEEE/CVF International Conference on Computer Vision, {(ICCV)}}, 2021,
  pp. 1728--1738.

\bibitem{radford2021learning}
A.~Radford, J.~W. Kim, C.~Hallacy, A.~Ramesh, G.~Goh, S.~Agarwal, G.~Sastry,
  A.~Askell, P.~Mishkin, J.~Clark \emph{et~al.}, ``Learning transferable visual
  models from natural language supervision,'' in \emph{International Conference
  on Machine Learning, {(ICML)}}, 2021, pp. 8748--8763.

\bibitem{lu2021learning}
W.~Lu, D.~Li, L.~Nie, P.~Jing, and Y.~Su, ``Learning dual low-rank
  representation for multi-label micro-video classification,'' \emph{IEEE
  Transactions on Multimedia}, 2021.

\bibitem{zhang2020hybrid}
Y.~Zhang, W.~Min, L.~Nie, and S.~Jiang, ``Hybrid-attention enhanced two-stream
  fusion network for video venue prediction,'' \emph{IEEE Transactions on
  Multimedia}, vol.~23, pp. 2917--2929, 2020.

\bibitem{qi2021semantics}
M.~Qi, J.~Qin, Y.~Yang, Y.~Wang, and J.~Luo, ``Semantics-aware spatial-temporal
  binaries for cross-modal video retrieval,'' \emph{IEEE Transactions on Image
  Processing}, vol.~30, pp. 2989--3004, 2021.

\bibitem{wang2022many}
W.~Wang, J.~Gao, X.~Yang, and C.~Xu, ``Many hands make light work: Transferring
  knowledge from auxiliary tasks for video-text retrieval,'' \emph{IEEE
  Transactions on Multimedia}, 2022.

\bibitem{qian2019video}
X.~Qian, Y.~Zhuang, Y.~Li, S.~Xiao, S.~Pu, and J.~Xiao, ``Video relation
  detection with spatio-temporal graph,'' in \emph{Proceedings of the ACM
  international conference on Multimedia}, 2019, pp. 84--93.

\bibitem{xiao2020visual}
J.~Xiao, X.~Shang, X.~Yang, S.~Tang, and T.-S. Chua, ``Visual relation
  grounding in videos,'' in \emph{European conference on computer vision,
  {(ECCV)}}, 2020, pp. 447--464.

\bibitem{li2021interventional}
Y.~Li, X.~Yang, X.~Shang, and T.-S. Chua, ``Interventional video relation
  detection,'' in \emph{Proceedings of the 29th ACM International Conference on
  Multimedia}, 2021, pp. 4091--4099.

\bibitem{wang2018videos}
X.~Wang and A.~Gupta, ``Videos as space-time region graphs,'' in
  \emph{Proceedings of the European Conference on Computer Vision, (ECCV)},
  2018, pp. 399--417.

\bibitem{yan2018spatial}
S.~Yan, Y.~Xiong, and D.~Lin, ``Spatial temporal graph convolutional networks
  for skeleton-based action recognition,'' in \emph{Proceedings of the AAAI
  Conference on Artificial Intelligence}, 2018.

\bibitem{wu2019learning}
J.~Wu, L.~Wang, L.~Wang, J.~Guo, and G.~Wu, ``Learning actor relation graphs
  for group activity recognition,'' in \emph{Proceedings of the IEEE/CVF
  Conference on Computer Vision and Pattern Recognition, {(CVPR)}}, 2019, pp.
  9964--9974.

\bibitem{kipf2016semi}
T.~N. Kipf and M.~Welling, ``Semi-supervised classification with graph
  convolutional networks,'' in \emph{International Conference on Learning
  Representations, {(ICLR)}}, 2017.

\bibitem{vaswani2017attention}
A.~Vaswani, N.~Shazeer, N.~Parmar, J.~Uszkoreit, L.~Jones, A.~N. Gomez,
  L.~Kaiser, and I.~Polosukhin, ``Attention is all you need,'' in
  \emph{Advances on Neural Information Processing Systems, {(NeurIPS)}}, 2017,
  pp. 6000--6010.

\bibitem{devlin2019bert}
J.~Devlin, M.-W. Chang, K.~Lee, and K.~Toutanova, ``Bert: Pre-training of deep
  bidirectional transformers for language understanding,'' in \emph{Proceedings
  of the 2019 Conference of the North American Chapter of the Association for
  Computational Linguistics: Human Language Technologies}, 2019, p.
  4171–4186.

\bibitem{ging2020coot}
S.~Ging, M.~Zolfaghari, H.~Pirsiavash, and T.~Brox, ``Coot: Cooperative
  hierarchical transformer for video-text representation learning,'' in
  \emph{Advances on Neural Information Processing Systems, {(NeurIPS)}}, 2020.

\bibitem{anderson2018bottom}
P.~Anderson, X.~He, C.~Buehler, D.~Teney, M.~Johnson, S.~Gould, and L.~Zhang,
  ``Bottom-up and top-down attention for image captioning and visual question
  answering,'' in \emph{Proceedings of the IEEE Conference on Computer Vision
  and Pattern Recognition, {(CVPR)}}, 2018, pp. 6077--6086.

\bibitem{hendrycks2016gaussian}
D.~Hendrycks and K.~Gimpel, ``Gaussian error linear units (gelus),''
  \emph{arXiv preprint arXiv:1606.08415}, 2016.

\bibitem{ba2016layer}
J.~L. Ba, J.~R. Kiros, and G.~E. Hinton, ``Layer normalization,'' \emph{arXiv
  preprint arXiv:1607.06450}, 2016.

\bibitem{xu2016msr}
J.~Xu, T.~Mei, T.~Yao, and Y.~Rui, ``Msr-vtt: A large video description dataset
  for bridging video and language,'' in \emph{Proceedings of the IEEE
  Conference on Computer Vision and Pattern Recognition, {(CVPR)}}, 2016, pp.
  5288--5296.

\bibitem{Yu_2018_ECCV}
Y.~Yu, J.~Kim, and G.~Kim, ``A joint sequence fusion model for video question
  answering and retrieval,'' in \emph{Proceedings of the European Conference on
  Computer Vision, (ECCV)}, 2018, pp. 487--503.

\bibitem{chen2011collecting}
D.~Chen and W.~B. Dolan, ``Collecting highly parallel data for paraphrase
  evaluation,'' in \emph{Proceedings of the 49th Annual Meeting of the
  Association for Computational Linguistics: Human Language Technologies},
  2011, pp. 190--200.

\bibitem{venugopalan2015sequence}
S.~Venugopalan, M.~Rohrbach, J.~Donahue, R.~Mooney, T.~Darrell, and K.~Saenko,
  ``Sequence to sequence-video to text,'' in \emph{Proceedings of the IEEE/CVF
  International Conference on Computer Vision, {(ICCV)}}, 2015, pp. 4534--4542.

\bibitem{zhou2018towards}
L.~Zhou, C.~Xu, and J.~J. Corso, ``Towards automatic learning of procedures
  from web instructional videos,'' in \emph{Proceedings of the AAAI Conference
  on Artificial Intelligence}, 2018.

\bibitem{russakovsky2015imagenet}
O.~Russakovsky, J.~Deng, H.~Su, J.~Krause, S.~Satheesh, S.~Ma, Z.~Huang,
  A.~Karpathy, A.~Khosla, M.~Bernstein \emph{et~al.}, ``Imagenet large scale
  visual recognition challenge,'' \emph{International journal of computer
  vision}, vol. 115, no.~3, pp. 211--252, 2015.

\bibitem{szegedy2017inception}
C.~Szegedy, S.~Ioffe, V.~Vanhoucke, and A.~Alemi, ``Inception-v4,
  inception-resnet and the impact of residual connections on learning,'' in
  \emph{Proceedings of the AAAI Conference on Artificial Intelligence},
  vol.~31, no.~1, 2017.

\bibitem{kay2017kinetics}
W.~Kay, J.~Carreira, K.~Simonyan, B.~Zhang, C.~Hillier, S.~Vijayanarasimhan,
  F.~Viola, T.~Green, T.~Back, P.~Natsev \emph{et~al.}, ``The kinetics human
  action video dataset,'' \emph{arXiv preprint arXiv:1705.06950}, 2017.

\bibitem{carreira2017quo}
J.~Carreira and A.~Zisserman, ``Quo vadis, action recognition? a new model and
  the kinetics dataset,'' in \emph{Proceedings of the IEEE Conference on
  Computer Vision and Pattern Recognition, {(CVPR)}}, 2017, pp. 6299--6308.

\bibitem{kingma2015adam}
D.~P. Kingma and J.~Ba, ``Adam: A method for stochastic optimization,'' in
  \emph{International Conference on Learning Representations, {(ICLR)}}, 2015.

\bibitem{bertasius2021space}
G.~Bertasius, H.~Wang, and L.~Torresani, ``Is space-time attention all you need
  for video understanding,'' \emph{arXiv preprint arXiv:2102.05095}, vol.~2,
  no.~3, p.~4, 2021.

\end{thebibliography}

\vfill


\end{document}